\documentclass[conference]{IEEEtran}

\usepackage[usenames,dvipsnames,table,xcdraw]{xcolor}
\usepackage{xspace}

\newcommand{\flair}{{FLAIR}\xspace}

\usepackage{times}
\usepackage{float}
\usepackage{amsfonts}
\usepackage{amsmath}
\usepackage{listings}
\usepackage{mdframed}
\usepackage{soul}
\usepackage{inconsolata}
\usepackage{tcolorbox}
\usepackage{fancyvrb}
\usepackage{caption}
\lstdefinestyle{customstyle}{
    language=TeX, 
    basicstyle=\small\ttfamily,
    keywordstyle=\color{blue},
    commentstyle=\color{green!50!black},
    numbers=left,
    numberstyle=\tiny,
    numbersep=5pt,
    breaklines=true,
    showstringspaces=false,
    frame=none,
}

\sethlcolor{orange!30}

\newenvironment{grayitemize}{%
  \begin{mdframed}[
    backgroundcolor=gray!10, 
    linewidth=0pt,           
    innerleftmargin=-5pt,     
    innerrightmargin=3pt,    
    innerbottommargin=5pt,    
    innertopmargin=5pt        
  ]
  \begin{itemize}
}{
  \end{itemize}
  \end{mdframed}
}

\usepackage{hyperref}
\usepackage{bbm}
\hypersetup{
    colorlinks=true,
    linkcolor=orange,
    filecolor=orange,      
    urlcolor=orange,
    citecolor=orange,
}
\usepackage[capitalise, nameinlink]{cleveref}

\usepackage[numbers, sort&compress]{natbib}
\usepackage{multicol}
\usepackage{graphicx}

\begin{document}


\title{FLAIR: Feeding via Long-horizon AcquIsition of Realistic dishes}













%
\author{\authorblockN{Rajat Kumar Jenamani\authorrefmark{1}$^1$,
Priya Sundaresan\authorrefmark{1}$^2$,
Maram Sakr$^3$,
Tapomayukh Bhattacharjee\authorrefmark{2}$^1$,
Dorsa Sadigh\authorrefmark{2}$^2$
}
\authorblockA{\authorrefmark{1}Equal Contribution, \authorrefmark{2}Equal Advising}
\authorblockA{$^1$Cornell University, $^2$Stanford University, $^3$University of British Columbia}
}


\twocolumn[{%
\renewcommand\twocolumn[1][]{#1}%
\maketitle

\vspace{-10pt}

\begin{center}
    \captionsetup{type=figure}
    \vspace{-0.3cm}
    \includegraphics[width=\textwidth]{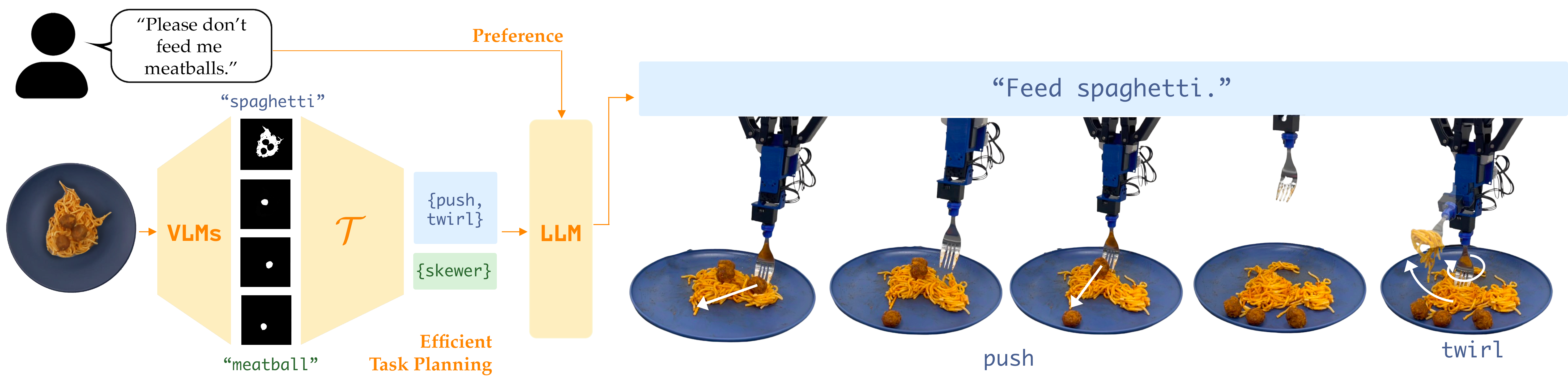}
    \captionof{figure}{We propose \flair, a system for long-horizon robot-assisted feeding that combines the commonsense and few-shot reasoning capabilities of foundation models  with a library of parameterized skills. Above, \flair takes visual observations and a given user preference (``Please don't feed me any meatballs'') to plan a sequence of actions that pushes aside meatballs and twirls spaghetti.}
    \label{fig:front}
\end{center}
}]

\begin{abstract}
Robot-assisted feeding has the potential to improve the quality of life for individuals with mobility limitations who are unable to feed themselves independently. However, there exists a large gap between the homogeneous, curated plates existing feeding systems can handle, and truly in-the-wild meals. Feeding realistic plates is immensely challenging due to the sheer range of food items that a robot may encounter, each requiring specialized manipulation strategies which must be sequenced over a long horizon to feed an entire meal. An assistive feeding system should not only be able to sequence different strategies \emph{efficiently} in order to feed an entire meal, but also be mindful of user \emph{preferences} given the personalized nature of the task. We address this with \flair, a system for long-horizon feeding which leverages the commonsense and few-shot reasoning capabilities of foundation models, along with a library of parameterized skills, to plan and execute user-preferred and efficient bite sequences. In real-world evaluations across 6 realistic plates, we find that \flair can effectively tap into a varied library of skills for efficient food pickup, while adhering to the diverse preferences of 42 participants without mobility limitations as evaluated in a user study. We demonstrate the seamless integration of \flair with existing bite transfer methods~\cite{gallenberger2019transfer, jenamani2024bitetransfer}, and deploy it across 2 institutions and 3 robots, illustrating its adaptability. Finally, we illustrate the real-world efficacy of our system by successfully feeding a care recipient with severe mobility limitations. Supplementary materials and videos can be found at: \href{http://emprise.cs.cornell.edu/flair}{emprise.cs.cornell.edu/flair}. 
\end{abstract}

\label{sec:intro}
\vspace{-0.3cm}
\section{Introduction}


Eating is a vital part of everyday life, yet millions worldwide struggle to feed themselves independently due to mobility limitations caused by conditions such as neurological disorders, injuries, the effects of aging, or other health complications~\cite{taylor2018americans}. These individuals often rely on caregivers for meal assistance, which impacts their sense of independence, daily routines, and the social experience of dining~\cite{jacobsson2000eatingprocess, nanavati2023design, shune2020experience}. Moreover, feeding is one of the most time-consuming Activities of Daily Living (ADL) for caregivers~\cite{chio2006caregiver}. A system for autonomous mealtime assistance holds promise for improving the quality of life for those requiring assistance~\cite{brose2010role}, and reducing the physical workload on caregivers~\cite{bhattacharjee2020moreautonomy, jenamani2024bitetransfer}. 

Robot-assisted feeding entails first performing \emph{bite acquisition}~\cite{gordon2020adaptive, gordon2021leveraging, gordon2023towards, sundaresan2022learning, sundaresan2023learning, feng2019robot, grannen2022learning, keely2024kiri}, where the robot must manipulate a utensil to pick up a 
bite of food, followed by \emph{bite transfer}~\cite{gallenberger2019transfer, park2020active, shaikewitz2022mouth, jenamani2024bitetransfer, belkhale2022balancing}, or bringing a bite of food to the mouth for consumption. 
In this paper, we primarily focus on bite acquisition. 
Several prior works in bite acquisition develop individual skills targeting specific food groups. This includes policies for skewering firm foods~\cite{feng2019robot, gordon2023towards, gordon2021leveraging, sundaresan2022learning, gordon2020adaptive}, scooping soft foods~\cite{grannen2022learning, tai2023scone}, or rearranging and twirling noodles~\cite{sundaresan2023learning}. These works, however, mostly operate over a single bite horizon or consider plates with a homogeneous type of food, such as only noodles or only bite-sized fruits and vegetables. However, the challenge of achieving reliable bite acquisition for dishes encountered \emph{in-the-wild}, which contain multiple different food types within the same meal and require strategic skill sequencing over many timesteps, persists. 

Consider a robot tasked with feeding a meal with a fruit appetizer---bananas, celery, and watermelon with chocolate sauce and ranch dressing---and spaghetti and meatballs for the main course. 
The robot must not only execute specialized strategies, such as cutting bananas, skewering fruits, dipping in sauces, and grouping and twirling noodles, but also infer how to sequence them over a \emph{long horizon}, considering: \\ 
\underline{Efficiency}: For the main course, if meatballs incidentally rest on top of the spaghetti, the robot should prioritize efficiency by serving the meatball first. This sequencing exposes the spaghetti for subsequent bites, avoiding the inefficiency of pushing the meatball aside to access the spaghetti initially. \\
\underline{User Preferences}: However, if the user prefers to not eat meatballs, the robot must adjust the bite sequence accordingly. \\
\underline{Commonsense Reasoning}: In the absence of explicit user preferences, the robot must employ commonsense reasoning to correctly order / combine bites for human-like feeding. For the appetizer, it should pair celery with ranch, bananas with chocolate, and feed watermelon standalone, reflecting typical food pairings. For the main course, it should vary the serving order between spaghetti and meatballs to avoid repetitions. \\
We desire a system that considers all these criteria to achieve long-horizon bite acquisition via a library of skills, and finally integrates with frameworks for bite transfer~\cite{park2020active, belkhale2022balancing, shaikewitz2022mouth, jenamani2024bitetransfer, gallenberger2019transfer} to effectively feed complete meals.

In this work, we introduce \flair(\underline{F}eeding via \underline{L}ong-horizon \underline{A}cqu\underline{I}sition of \underline{R}ealistic dishes), a robot-assisted feeding system capable of feeding a complete meal to a care recipient. Given a plate image, and an optional user-provided natural language preference specifying their desired feeding strategy (i.e. `I prefer to alternate bites of X and Y' or `Don't feed me X'), \flair executes a sequence of actions that efficiently feeds the items on the plate while adhering to the preference. The framework starts by detecting food items and their semantic labels (i.e. $\texttt{`spaghetti'}$) via Vision-Language Models (VLMs). We then pass the visual state estimate and semantic label for all items to a hierarchical task planner, which outputs per-item efficiencies by proxy of inferring a sequence of skills to achieve acquisition for each item. Finally, we pass all of this context -- the food item labels, the optional user's preference, and per-item efficiences -- to a Large Language Model (LLM)-based planner which outputs the next bite to feed. The few-shot reasoning capabilities of LLMs allows for reasoning about the available context in a chain-of-thought manner, and planning sequences of bites that cater to both preference and efficiency. We carry out these action sequences via a library of parameterized food manipulation skills implemented on custom hardware. Finally, \flair's modular approach to long-horizon bite acquisition enables seamless integration with existing outside-mouth bite transfer~\cite{gallenberger2019transfer} and inside-mouth bite transfer~\cite{jenamani2024bitetransfer} frameworks. 

We deploy \flair across two institutions and three robots: a Kinova 6-DoF at Cornell University and a Franka Emika Panda and a Kinova 7-DoF at Stanford University, demonstrating its adaptability to various robotic platforms. We validate FLAIR for long-horizon food pickup across six diverse plates, ranging from DoorDash orders and prepared grocery store meals to homemade meals. In a user study across 42 individuals without mobility limitations, we use \flair to demonstrate the necessity of balancing between both preferences and efficiency for feeding complete, realistic meals, as compared to an efficiency-only or preference-only approach. Moreover, we compare \flair's hierarchical task planner against three state-of-the-art baselines~\cite{sundaresan2023learning,achiam2023gpt,liu2021swin} on two different datasets, demonstrating that it significantly outperforms these baselines. Finally, we demonstrate the real-world effectiveness of our system in feeding a care recipient with Multiple Sclerosis a meal consisting of various fruits and dips. 

Overall, our contributions include: 
\begin{itemize}
    \item \flair: A system for long-horizon feeding which
leverages foundation models to sequence a library of diverse skills towards in-the-wild long-horizon bite acquisition.
    \item Deployment of \flair across two institutions and three different robots, demonstrating its versatility.
    \item A user study with 42 individuals without mobility limitations across 6 diverse plates validating the effectiveness of considering both preferences and efficiency for feeding.
    \item Demonstration of the real-world efficacy of our system by feeding a care recipient with mobility limitations.
\end{itemize}
\section{Related Work}
\smallskip
\noindent \textbf{Robot-Assisted Feeding.} 
While various commercial robot-assisted feeding systems~\cite{obi, neater} have been introduced, they typically rely on pre-programmed trajectories or user teleoperation. This limited autonomy has hindered their widespread adoption and retention, and inspired autonomous methods for bite acquisition and transfer. Prior work in bite acquisition has focused on developing individual food manipulation skills for specific food types. Various works \cite{feng2019robot, gordon2020adaptive, gordon2021leveraging, sundaresan2022learning} tackle acquisition of solid bite-sized foods, and demonstrate effective skewering strategies based on the food item's pose and material properties.
Sundaresan et al. \cite{sundaresan2023learning} propose visually parameterized primitives for twirling and grouping noodle-like dishes, and show generalization to unseen noodles. Beyond fork-based manipulation, Grannen et al. \cite{grannen2022learning} plan bimanual scooping actions with two custom utensils, while Tai et al. \cite{tai2023scone} and Zhang et al. \cite{zhang2019leveraging} develop specialized strategies for scooping with a spoon and cutting with a knife, respectively. 
However, no prior work in robot-assisted feeding considers complete, in-the-wild  meals containing various food types (noodles, semisolids, sauces, cuttable food items, etc.) within the same plate, as typically encountered in everyday scenarios. 

In this work, we leverage insights from the aforementioned state-of-the-art food manipulation works to develop a large library of bite acquisition skills, and use foundation models to sequence these skills for efficiently feeding realistic dishes while obeying user preferences. To the best of our knowledge, FLAIR is the first of any autonomous feeding system to tackle in-the-wild meals containing various food types, and incorporate bite sequencing preferences for long-horizon feeding. 

Various works have shown joint bite acquisition with transfer~\cite{gallenberger2019transfer, bhattacharjee2020moreautonomy, jenamani2024bitetransfer}. However, they typically consider bite acquisition actions over a single timestep and not over the complete meal. In contrast, we illustrate that our long-horizon bite acquisition framework can seamlessly integrate with existing methods for bite transfer~\cite{jenamani2024bitetransfer,gallenberger2019transfer}, and demonstrate feeding of a full meal to a care recipient.


\smallskip
\noindent \textbf{Foundation Models for Robotic Manipulation.} 
Two of the most challenging aspects associated with feeding are planning over available skills, and developing a library of food manipulation skills themselves. To this end, several recent works in robotic manipulation use foundation models such as vision-language models (VLMs)~\cite{achiam2023gpt, karamcheti2023language,radford2021learning,li2022blip} or large language models (LLMs)~\cite{radford2019language, devlin2018bert,radford2018improving, brown2020language, chowdhery2023palm} towards both high-level task planning and skill instantiation. A standard approach is to prompt foundation models with context including available skills, object states, etc., and to use them to plan action sequences either for long-horizon manipulation~\cite{chen2023autotamp, wu2023tidybot, huang2022inner,agia2022taps} or grounded exploration~\cite{jiang2024roboexp}. However, these works focus on exploiting the commonsense reasoning capabilities of these models~\cite{kwon2023toward, sap2020commonsense, gao2023physically}, such as inferring a sequence of skills that is feasible or user-preferred based on the provided visual and semantic context. In the setting of feeding, we additionally care about planning \emph{efficient} skill sequences. Reasoning about efficiency and skill affordances in a few-shot manner remains brittle and challenging for these models, due to hallucinations~\cite{kambhampati2024can} and a lack of priors about embodied agents. Instead, we propose inferring the efficiencies of skills separately, and providing this as additional context to aid in planning.

Besides skill sequencing, several recent works show the benefits of using foundation models towards inferring the parameters of low-level skills themselves, rather than data-driven approaches to learning skill policies from scratch~\cite{brohan2022rt, brohan2023rt, padalkar2023open}. Recent approaches include instantiating skills via code skeletons generated by LLMs~\cite{liang2023code, huang2023voxposer}, or implementing skills parameterized by open-vocabulary object detectors~\cite{, zha2023distilling, wu2023tidybot} or keypoint affordances from VLMs~\cite{liu2024moka}. These approaches have mainly been applied to simple quasi-static actions such as pick and place. We instead apply this paradigm towards estimating the visual state of food items, and using this to parameterize a diverse library of skills such as twirling, scooping, and cutting. 

\smallskip
\noindent \textbf{User Preferences in Assistive Robotics.} The inclusion of user preferences in the design and operation of assistive robots is essential for significantly enhancing user satisfaction~\cite{canal2021preferences}. These preferences can be identified either implicitly through data-driven methods~\cite{yang2021desire} or explicitly stated by users~\cite{canal2019adapting}. Canal et al.~\cite{canal2019adapting} explore task planning adhering to user preferences for an assistive shoe dressing experiment. However, they explore user specification only in form of post-hoc scoring of executed actions which is restrictive for various safety critical applications. Madan et al. \cite{madan2022sparcs} propose training a hidden Markov Model with user-demonstrated sequencing data for the same meal collected over multiple days to learn preferred bite sequences, enhancing user satisfaction. However, this proof of concept did not involve a robot-assisted feeding system and is impractical to extend to the diverse meals an individual might consume. Recently, TidyBot~\cite{wu2023tidybot} showcased that LLMs can summarize information from limited examples and extrapolate general user preferences for determining the proper place to put each object while tidying a room. However, their approach to task planning lacks consideration of additional metrics, such as efficiency, which is crucial in our context of feeding a complete meal. \flair~instead factors in both user preference and acquisition efficiency for long-horizon feeding.
\section{FLAIR: \underline{F}eeding via \underline{L}ong-horizon \underline{A}cquis\underline{i}tion of 
\underline{R}ealistic dishes}
In this section, we present \flair, a system for feeding complete meals which combines existing foundation models in a novel way towards personalized and efficient bite sequencing. We first give an overview of our custom system hardware, then outline our approach to long-horizon bite acquisition, and finally discuss integration of our method with existing bite transfer frameworks~\cite{gallenberger2019transfer, jenamani2024bitetransfer} for feeding of in-the-wild dishes. 

\begin{figure}
    \centering
\includegraphics[width=0.5\textwidth]{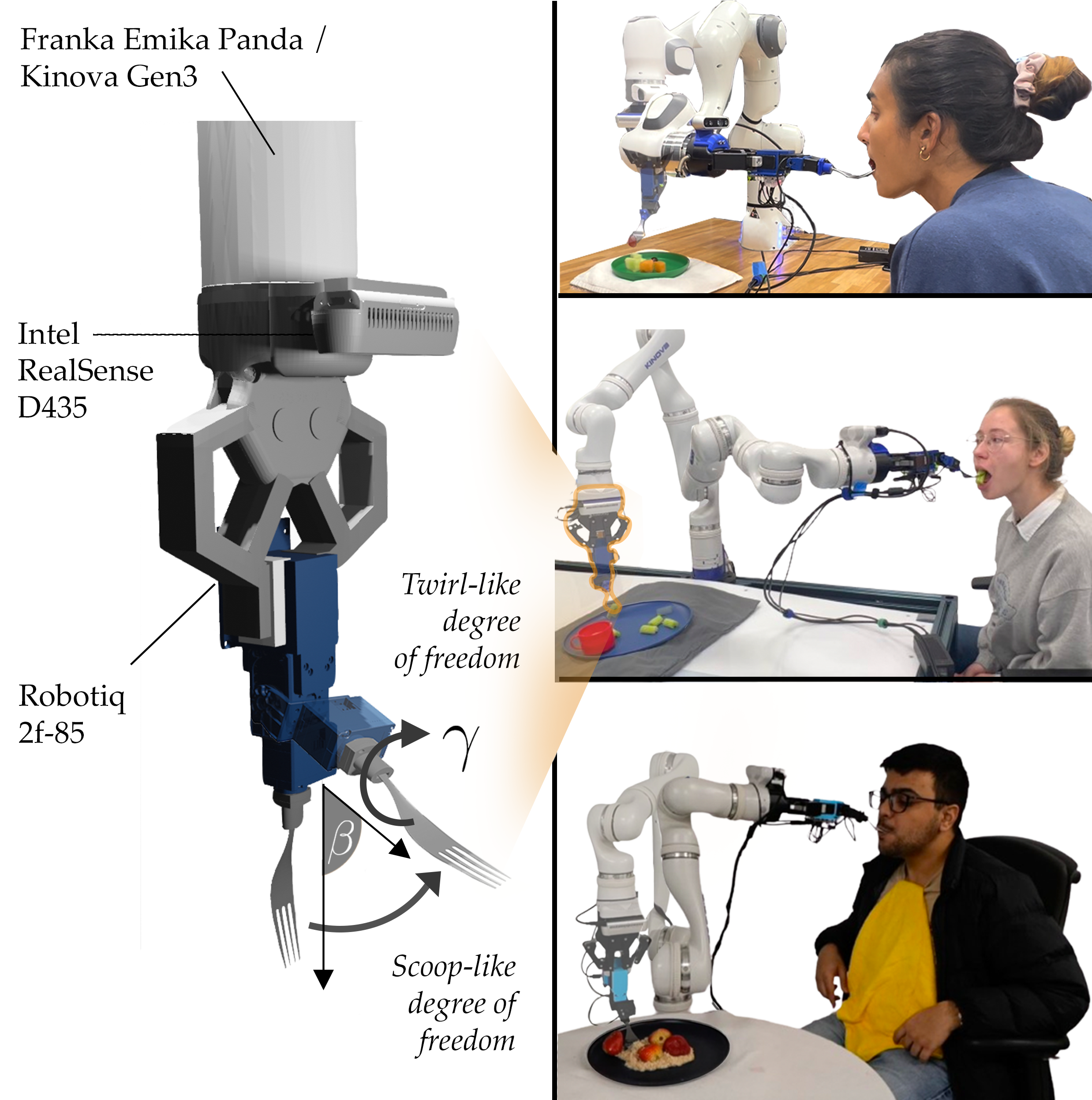}
    \caption{We implement our skill library using a custom feeding utensil (adapted from~\cite{shaikewitz2022mouth}) having two degrees of freedom for easy twirling and scooping at the end effector. We deploy the full feeding stack on three robots and two institutions: the 7-DoF Franka Emika Panda (top) and 7-DoF Kinova Gen 3 (middle) at Stanford University, and the 6-DoF Kinova Gen 3 (bottom) at Cornell University.} 
    \label{fig:hardware_setup}
    \vspace{-15pt}
\end{figure}

\subsection{Hardware System}
We tackle a wide range of food categories in this work such as fruits, vegetables, noodles, meat, soft foods, dipping sauces, and non-bite-sized items that require cutting. Many of these foods require specialized, dynamic manipulation strategies that typical 6 or 7-DoF robots struggle with due to their limited workspace. We thus implement \flair~on Kinova and Franka robot arms equipped with a motorized feeding utensil mounted at the end-effector, adapting the design from~\cite{shaikewitz2022mouth}. The utensil contains a fork attachment and has two degrees of freedom corresponding to the orientation of the fork tines and the tilt angle. This allows for directly controlling the utensil to perform dynamic movements like twirling and scooping, while the robot handles moving between waypoints in the workspace via Cartesian position control. We also use a wrist-mounted RGB-D Realsense camera with a known end-effector to camera transformation. This enables perceiving plates of food and localizing food items in the 3D workspace. We note that the same hardware was replicated on two different Kinova arms and one Franka Emika Panda, each with their separate fork attachment and sensors across two different institutions (detailed in Appendix), demonstrating the reproducibilty of our method and hardware (\cref{fig:hardware_setup}).

\subsection{Long-Horizon Bite Acquisition Framework}
\label{sec:bite_acq}
With access to a hardware platform that supports dexterous food manipulation strategies, our goal is to plan and execute long-horizon bite sequences that cater to a user's preference while efficiently feeding a meal.

\smallskip
\noindent \textbf{Problem Formulation.}
\label{sec:ps}
 We assume access to an RGB-D plate image observation $o_t \in \mathcal{O} = \mathbb{R}_{+}^{W \times H \times 4}$ of width $W$ and height $H$, and an optional natural language instruction $\ell_{pref}$ from the user, representing their preferred feeding strategy at a high-level (i.e., $\ell_{pref}$ = ``Feed me alternating bites of X and Y'' or ``Only feed me X''). X and Y can denote an arbitrary food item semantic label (i.e. ``spaghetti'', ``strawberry'', ``caramel'') or category (i.e. ``noodles'', ``fruit'', ``sauce''). 

We further assume access to a library $\mathcal{L} = \{\phi^1, \hdots, \phi^N\}$ of $N$ skills that the robot can use to manipulate food items. Each skill $\phi^i(p)$ represents a parameterized manipulation primitive that takes in parameters $p$ and outputs low-level motor commands. We represent a low-level action at time $t$ by $a_t = (x,y,z,\beta,\gamma,\psi)$, where $(x,y,z)$ denotes the position of the feeding utensil tip, $\beta$ and $\gamma$ denote pitch and roll of the utensil respectively, and $\psi$ denotes the robot's end effector roll angle. Thus, the output of any skill is a sequence of $T$ actions $\{a_t, a_{t +1}, \hdots, a_{t + T}\}$ that the robot takes to execute the particular strategy. For instance, a skewering skill may take the position and orientation of a desired food item as input, and output a trajectory that skewers the item of choice. Our goal is to plan and execute a sequence of parameterized skills $\{\phi_1(p_1), \phi_2( p_2), \hdots, \phi_H(p_H)\}$ which results in efficient and user-preferred bite acquisition, where $H$ is the total number of skills to execute to complete feeding a plate and $p_h$ refers to the parameters of skill $\phi_h \in \mathcal{L}$.

\begin{figure*}
    \centering
    \makebox[\textwidth][c]{%
    \includegraphics[width=1\textwidth]{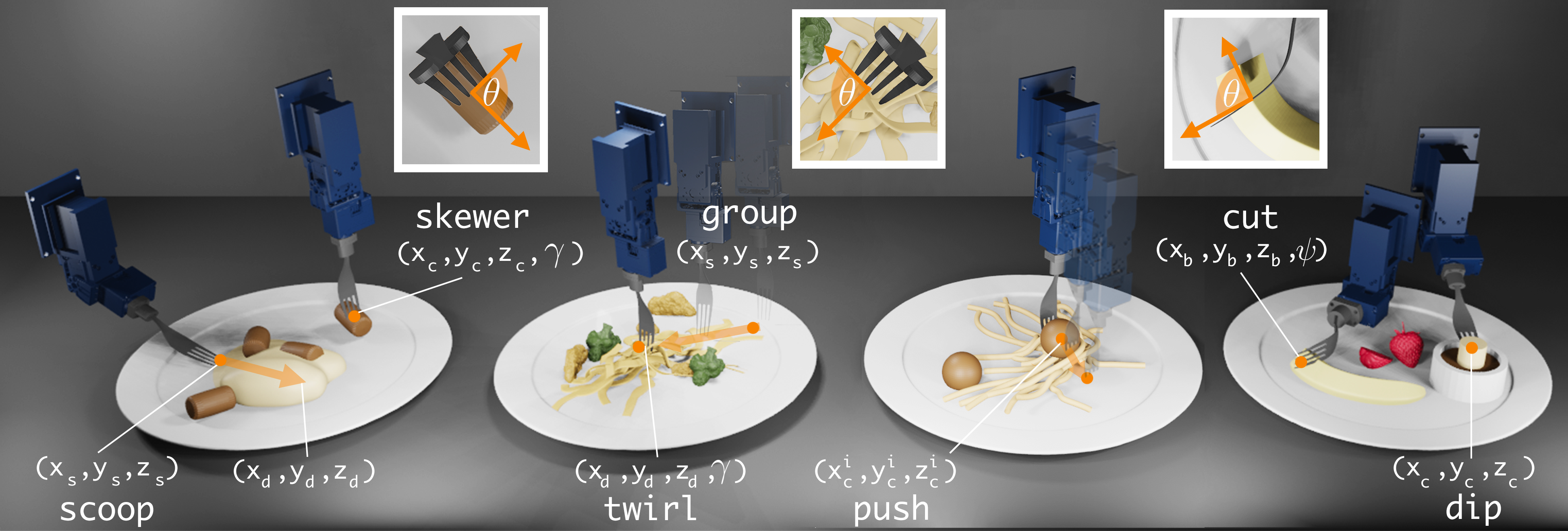}
    }
    \caption{Our skill library consists of 7 parameterized manipulation skills: 4 acquisition (skewer, twirl, scoop, dip) and 3 pre-acquisition (group, push, cut).}
    \label{fig:skills}
    \vspace{-15pt}
\end{figure*}

\smallskip
\noindent \textbf{State Representations for Food.}
Our approach addresses the main challenges in long-horizon bite acquisition—parameterizing low-level skills and sequencing them—by integrating state-of-the-art visual-language models. We use visual state estimates and semantic features of food items to guide skill parameterization and sequencing.

For a given plate observation $o_t$ at time t, we first query GPT-4V~\cite{achiam2023gpt} in a few-shot manner to recognize which food items are present. We prompt the model with a few in-context examples of plate images and their corresponding ground truth food item semantic labels, and ask the model to complete the prompt for the test image $o_t$. GPT-4V outputs a list of semantic labels $l_t$ that are present, (i.e., $l_t = \texttt{[`fettuccine', 'chicken', `broccoli']}$) along with their corresponding categories $c_t$ (i.e., $c_t = [\texttt{`noodles'}, \texttt{`meat/seafood'}, \texttt{`vegetable'}, \texttt{`cuttable'}]$). These categories are relevant for associating the appropriate skill to each food item for bite acquisition. We then pass the recognized semantic labels to GroundingDINO~\cite{liu2023grounding}, an open-vocabulary VLM, for bounding box detection. For each bounding box, we use SegmentAnything (SAM)~\cite{kirillov2023segment} to refine these bounding boxes into segmentation masks $\{m^1_t, m^2_t, \hdots, m^D_t\}$ for all $D$ items detected. 



\smallskip
\noindent \textbf{Skill Library}
The segmented representations of food we obtain from VLMs provide a useful way to parameterize food manipulation skills, which we split into \emph{acquisition} and \emph{pre-acquisition} skills. \cref{fig:skills} visualizes all skill parameterizations.

\smallskip
\subsubsection{Acquisition skills} Acquisition skills refer to those that pick up food, such as skewering a food item, twirling a pile of noodles, scooping a soft pile of food, or dipping an item to coat it in sauce. We parameterize them as follows, assuming access to a segmentation mask $m^i_t$ for the item of interest:
\begin{grayitemize}

    \item \underline{\hl{\texttt{skewer}$(x_c, y_c, z_c,\gamma)$}}: We detect the centroid of $m^i_t$ and deproject this 2D pixel coordinate to a 3D coordinate $(x_c, y_c, z_c)$ representing the center of a food item in the robot's frame of reference. We also estimate the major axis orientation $\theta$ of an item from $m^i_t$ analytically. Following~\cite{feng2019robot, sundaresan2022learning}, we bring the utensil above the the food item center with $\gamma = 90^{\circ} + \theta$ and execute a swift downward trajectory skewering perpendicular to the main axis of the item. This encourages the tines of the fork to pierce the item. If the tines align parallel to the item's major axis, they may run along its longer length and miss the shorter breadth due to slight calibration challenges, leading to unsuccessful skewering. 
    \item \underline{\hl{\texttt{twirl}$(x_d, y_d, z_d,\gamma)$}}: We adopt the parameterization from VAPORS~\cite{sundaresan2023learning}, a long-horizon system for noodle acquisition.  Specifically, we twirl noodles by bringing the fork to the sensed \emph{densest} pile $(x_d, y_d, z_d)$  on the plate, estimated via 2D Gaussian filtering on $m^i_t$, and with $\gamma$ identical to the parameterization for skewering (orthogonal to the major axis of the noodle pile sensed via a pose estimation network from~\cite{sundaresan2023learning}). We actuate the roll joint of the fork to complete two full twirls, wrapping noodles on the fork.
    \item \underline{\hl{\texttt{scoop}$(x_s, y_s, z_s, x_d, y_d, z_d)$}}: The fork starts with tines horizontal to the plate and scoops from the \emph{sparsest} region $(x_s, y_s, z_s)$ to the densest region $(x_d, y_d, z_d)$ on the plate, up to a pre-defined maximum distance empirically selected to pick up a bite-sized amount. We define the sparsest region as the point on the boundary of the food item mask $m^i_t$ that is furthest from the densest region, with the condition that the line connecting these points is not intersected by other food items, such as toppings.
    \item \underline{\hl{\texttt{dip}$(x_c, y_c, z_c)$}}: Finally, dipping entails bringing a fork containing a food item into the center $(x_c, y_c, z_c)$ of a small dish containing sauce. We initially orient the fork with tines horizontal to the plate to avoid the food item slipping off the utensil during dipping.
\end{grayitemize}
Immediately following each of these actions, the robot moves the fork tines in a scooping motion by actuating the utensil's pitch joint. The resulting horizontal fork helps prevent items from slipping off the fork after being picked up.

\subsubsection{Pre-acquisition skills} When the above acquisition skills are not immediately feasible due to occlusion from other items or the anticipated amount of food to be picked up being insufficient, we employ a number of auxiliary strategies which we refer to as \emph{pre-acquisition} skills. 
These actions do not directly pick up food but rearrange or manipulate items to facilitate subsequent acquisition. Grouping noodles into a pile before twirling, pushing a meatball off of a bed of spaghetti before twirling, or cutting banana into a bite-sized piece before pickup are all examples of pre-acquisition. We parameterize them as follows:

\begin{grayitemize}
    \item \underline{\hl{\texttt{group}$(x_s, y_s, z_s, x_d, y_d, z_d)$}}: For a pile of food distributed on the plate, we sense the densest $(x_d, y_d, z_d)$ and sparsest $(x_s, y_s, z_s)$ regions via $m^i_t$ and execute a linear push with the fork from the sparsest to densest point.
    \item \underline{\hl{\texttt{push}$(x^i_c, y^i_c, z^i_c)$}}: For a food item with mask $m^i_t$ obstructing a pile of food (such as noodles or a semisolid) with mask $m^j_t$, we can execute a linear push motion starting at the centroid of the obstructing item, to the nearest boundary point of the underlying food bed $m^j_t$.
    \item \underline{\hl{\texttt{cut}$(x_b, y_b, z_b, \psi)$}}: To cut a food item, we estimate a point on the object $(x_b, y_b, z_b)$ that would result in a bite-sized portion once cut. In practice, we detect the major axis of $m^i_t$ and traverse a fixed unit length from the one end of the axis to estimate this. We then bring the fork horizontal ($\beta = 90^\circ$) and with sideways tines ($\gamma = 90^\circ$). Finally, we set the end-effector roll $\psi$ such that the lateral side of the fork is orthogonal to the major axis angle $\theta$ as in skewering and twirling. Then, we execute a swift downward trajectory to slice the soft item.
    
\end{grayitemize}


$\mathcal{L} = \{\texttt{skewer}, \texttt{twirl}, \texttt{scoop}, \texttt{dip}, \texttt{group}, \texttt{push}, \texttt{cut}\}$ forms the library of vision-parameterized skills at the core of \flair. 

We provide further details on the vision-based parameterizations for each of these skills in the Appendix.

\smallskip
\noindent \textbf{Task Planning for Acquisition.} We plan a sequence of bites that both satisfies the preference of the user, and is efficient for the robot to acquire. The latter consideration requires reasoning over the sequence of pre-acquisition and acquisition skills needed to pick up an item, for which we introduce a hierarchical task planner $\mathcal{T}$. Our task planner relies on vision modules which post-process the segmented plate observations to quantify the density and spread of food items, along with checking for appropriate bite sizes and collision with other food items.  While it uses a few key parameters, the overall pipeline and these parameters are shared across different categories like $\texttt{`noodles'}$, $\texttt{`semisolid'}$, and $\texttt{`cuttable'}$, rendering the approach versatile for diverse plates. 


The task planner takes as input a particular food item category $c^i_t$ along with the detected segmentation mask $m^i_t$ and outputs a sequence of skills to acquire the item. The skill library in this work addresses the following categories of food items: $\{\texttt{`meat/seafood'}, \texttt{`fruit'}, \texttt{`vegetable'}, \texttt{`sauce'},\\ \texttt{`noodles'},\texttt{`semisolid'},  \texttt{`cuttable'} \}$. 

For most categories, acquisition tends to be immediately possible. Food items such as a bite of $\{\texttt{`meat/seafood'},$ $ \texttt{`fruit'}, \texttt{`vegetable'}\}$ tend to be isolated on a plate and immediately acquirable. Thus, we plan the following acquisition skills, where $p^i_t$ denotes the parameters of the skill to manipulate the $i$-th food item, sensed from $m^i_t$ and $o_t$:

\begin{grayitemize}
    \item $\mathcal{T}(c^i_t, m^i_t) = \{\texttt{skewer}(p^i_t)\}$ for \\ $c^i_t \in \{\texttt{`meat/seafood'}, \texttt{`fruit'}, \texttt{`vegetable'}\}$ 
    \item $\mathcal{T}(\texttt{`sauce'}, m^i_t) = \{\texttt{dip}(p^i_t)\}$
\end{grayitemize}

 Food items that are instead in the $\texttt{`noodles'}$, $\texttt{`semisolid'}$, or $\texttt{`cuttable'}$ category require more nuanced reasoning about pre-acquisition depending on the distribution of the food on the plate, and whether other food items are intermixed, on top, or to the side. 
 We critically observe that the segmentation mask $m^i_t$ obtained from the VLM provides a useful prior over the spread of food on the plate, which can guide action selection. 
 We apply a Gaussian smoothing kernel over $m^i_t$ which has the effect of producing a normalized density heatmap of the food,
 and use simple pre-conditions to determine a sequence of skills to pick up a bite of noodles or a semisolid.
Specifically, we measure the maximum \emph{density} and the 2D \emph{entropy} of the heatmap and plan actions as follows.

If the density exceeds a pre-defined threshold $\texttt{DENSITY\_THRESH}$, this indicates the presence of a large pile of food that can be immediately acquired:
\begin{grayitemize}
    \item  $\mathcal{T}(\texttt{`noodles'}, m^i_t) = \{\texttt{twirl}(p^i_t)\}$
    \item $\mathcal{T}(\texttt{`semisolid'}, m^i_t) = \{\texttt{scoop}(p^i_t)\}$.
\end{grayitemize}

However, when twirling is obstructed by another item mask $m^j_t$, such as a meatball too close to a planned spaghetti twirling action, the obstructing food must be pushed aside first. Similarly, if toppings such as sausages block all viable scooping actions for mashed potatoes, the sausage nearest to the boundary of the mashed potato mask should be pushed.

\begin{grayitemize}
    \item $\mathcal{T}(\texttt{`noodles'}, m^i_t) = \{\texttt{push}(p^i_{t,\texttt{push}}), \texttt{twirl}(p^i_{t,\texttt{twirl}})\}$ if the robot must push aside a topping before twirling.
    \item $\mathcal{T}(\texttt{`semisolid'}, m^i_t) = \{\texttt{push}(p^i_{t,\texttt{push}}), \texttt{scoop}(p^i_{t,\texttt{scoop}})\}$ if the robot must push aside a topping before scooping.
\end{grayitemize}

For food items in the $\texttt{`noodles'}$ category, we also consider grouping actions. If the entropy exceeds a pre-defined threshold $\texttt{ENTROPY\_THRESH}$, indicating that the food item is spread out on the plate, grouping can be helpful. We can directly execute unobstructed grouping actions. However, if all viable grouping actions are blocked by toppings, we instead push the topping closest to the boundary of the underlying food aside, group the food, and then acquire.

\begin{grayitemize}
    \item  $\mathcal{T}(\texttt{`noodles'}, m^i_t) = \{\texttt{group}(p^i_{t,\texttt{group}}), \texttt{twirl}(p^i_{t,\texttt{twirl}})\}$ if grouping is unobstructed.
    \item $\mathcal{T}(\texttt{`noodles'}, m^i_t) = \{\texttt{push}(p^i_{t,\texttt{push}}), \texttt{group}(p^i_{t,\texttt{group}}), \\ \texttt{twirl}(p^i_{t,\texttt{twirl}})\}$ if the robot must push aside an obstructing topping before grouping.
\end{grayitemize}

If neither acquisition nor grouping is feasible according to the set thresholds, we push the topping within the food item mask nearest to its boundary, hoping to expose more of the food item for future acquisition. If no such topping is available, we default to acquisition.   

Finally, for $\texttt{cuttable}$ items like cake, we use a max major axis length threshold to determine if $m^i_t$ is bite-sized or not, and either cut and then skewer, or skewer immediately:
\begin{grayitemize}
    \item $\mathcal{T}(\texttt{`cuttable'}, m^i_t) = \{\texttt{cut}(p^i_{t,\texttt{cut}}), \texttt{skewer}(p^i_{t,\texttt{skewer}})\}$, if the length of major axis of $m^i_t$ exceeds $\texttt{BITE\_LENGTH}$,
    \item $\mathcal{T}(\texttt{`cuttable'}, m^i_t) = \{\texttt{skewer}(p^i_t)\}$, otherwise.
\end{grayitemize}

We provide further details on task planning for acquisition in the Appendix. 

\subsection{Bite Sequencing via Foundation Models}
We introduce a unified framework for planning and executing bite sequences that are efficient and adhere to user preferences. With access to a library of skills $\mathcal{L}$, task planner $\mathcal{T}$, and user preference $\ell_{pref}$, we show how the commonsense-reasoning capabilities of LLMs enable them to act as few-shot planners for bite sequencing, inherently balancing preference and efficiency.

We prompt an LLM, in our case GPT-4V with relevant context about the meal. This includes the semantic food item labels $l_t$ and the user's preference $\ell_{pref}$. We augment this context with: (i) a \emph{history} of bites taken so far, (ii) an estimate of the \emph{portions} of each food type remaining, and (iii) the \emph{per-item efficiencies} which correspond to the number of actions required to pick up a food item (i.e. $|\mathcal{T}(c^i_t, m^i_t)|$). A higher number indicates a less efficient bite option since more pre-acquisition skills are required to pick up the item. To estimate quantities, we simply count the number of instances detected of the food item present if the item category is $[\texttt{`fruit'}, \texttt{`meat/seafood'}, \texttt{`vegetable'}]$. For $\texttt{`cuttable'}$ food items, we use the major axis length of $m^i_t$ divided by $\texttt{BITE\_LENGTH}$. Otherwise for food items where the ground truth quantity is not countable, such as for $\texttt{`noodles'}$ or $\texttt{`semisolid'}$, we use $\lceil |m^i_t|/\texttt{PORTION\_SIZE}\rceil$, where $\texttt{PORTION\_SIZE}$ is empirically determined for approximating the number of bite portions in a mask.

Below, we show an example input prompt in gray with the immediate next bite planned by asking the LLM to perform completion. We abridge the prompt here for brevity, but include our full prompting strategy in Appendix.

\begin{tcolorbox}[colback=gray!10, colframe=gray!40, boxrule=0.1pt, sharp corners, fontupper=\small, left=0pt,]
\begin{verbatim}
Items remaining: ["fettuccine", "chicken", "broccoli"]
Preference: "Alternating bites of each"
History: ["chicken"]
Portions remaining: [5, 1, 2]
Efficiencies: [3, 1, 1]
---
Decide what bite to feed me next.
Format your response as follows:
Strategy: Sentence describing your high-level strategy
Next bite: Phrase describing the next bite you will feed
Next bite as list: ['item'] # Or ['item', 'dip'] or []
---
Output your response here.
\end{verbatim}
\end{tcolorbox}
\begin{tcolorbox}[colback=orange!10, colframe=gray!40, boxrule=00pt, sharp corners, fontupper=\small, left=0pt,]
\begin{verbatim}
Strategy: Given that you want to alternate amongst the
three types of items present, and you just ate chicken,
I will feed you either fettucine or broccoli.
Even though broccoli is a more efficient option, with only
1 action required, I will feed you fettuccine since 
there are more portions of it.
Next bite: Feed fettuccine.
Next bite as list: ['fettuccine']
\end{verbatim}
\end{tcolorbox}

By reasoning about the provided context in a chain-of-thought style, the LLM generates a subsequent bite. Importantly, there are no explicit tradeoffs between efficiency, portion size, or preference which we embed into the prompt, allowing the LLM to reason about the most sensible strategy based on the intensity of the user's preference and available plate context. Given a next bite, in this case $\texttt{`fettuccine'}$, we then plan the appropriate skill sequence via $\mathcal{T}(\texttt{`noodles'}, m^{\texttt{`fettuccine'}}_t$) and execute it via $\mathcal{L}$.

\subsection{Integration of Acquisition and Transfer}
The self-contained nature of our \textit{bite acquisition} framework allows for straightforward integration with \emph{bite transfer} frameworks, and is agnostic to the exact approach used. We demonstrate this easy combination with two existing methods: (i) an outside-mouth bite transfer method~\cite{gallenberger2019transfer} that features visual servoing capabilities, and (ii) a recent method for inside-mouth transfer~\cite{jenamani2024bitetransfer} that leverages robust mouth tracking and physical interaction-aware control.

A significant challenge in this integration is ensuring that food, particularly semi-solid items such as mashed potatoes or noodle-like items such as spaghetti, does not spill while it moves from above the plate to the pre-transfer pose infront of the mouth. Prior works with non-actuated utensils \cite{park2020active} use an MPC-based approach to generate robot trajectories that constrain the orientation of the utensil to remain upright. However, these methods often require complex tuning and can be prone to getting trapped in local minima. Our feeding utensil (\cref{fig:hardware_setup}) enables us to uniquely circumvent this challenge. We leverage its roll ($\gamma$) and pitch ($\beta$) degrees of freedom, distinct from the robot's own degrees of freedom, to consistently keep the fork's tines horizontal regardless of the robot's motion. We continuously monitor the robot's end-effector pose at 10 Hz and adjust the feeding utensil's joints accordingly, ensuring a smooth and spill-free transfer of food to the user's mouth.


\begin{figure*}
    \centering
    \includegraphics[width=1.0\textwidth]{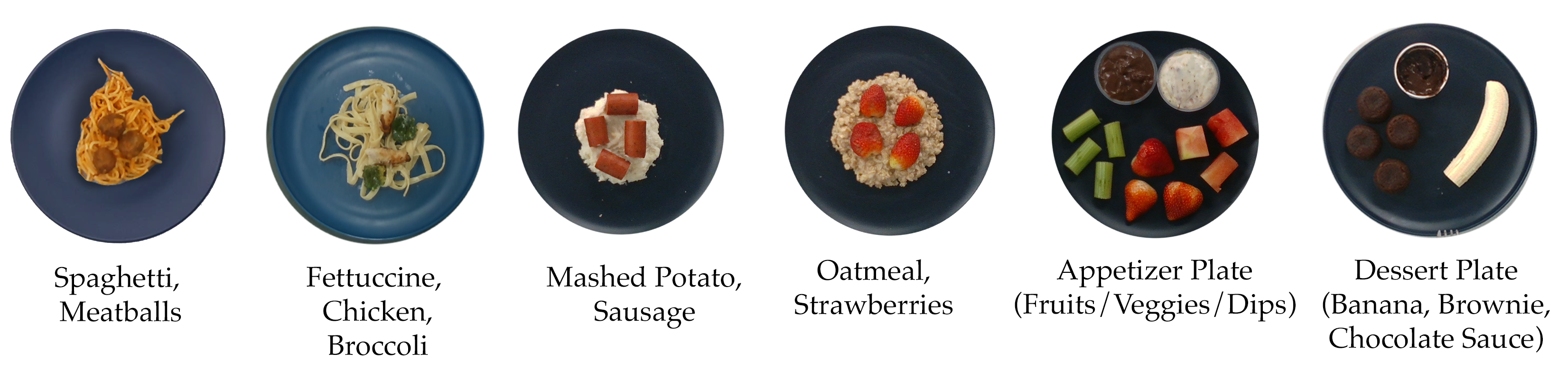}
    \vspace{-20pt}
    \caption{\textbf{Plates}: We evaluate our system on the following six plates containing a variety of food items, each necessitating highly different manipulation skills.} 
    \label{fig:plates}
    \vspace{-5pt}
\end{figure*}
\vspace{-0.05cm}
\section{Experiments}

We evaluate the effectiveness of \flair~ for feeding diverse plates each containing various types of food items. We first conduct a user study to assess \flair's ability to perform long-horizon bite acquisition of in-the-wild plates, while adhering to user preferences and efficiently feeding bites. For all acquisition experiments, we interchangeably use 2 Kinova Gen3 arms (one 6-DoF, and another 7-DoF), and a 7-DoF Franka Emika Panda. We then ablate our hierarchical task planner $\mathcal{T}$ against various state-of-the-art baselines~\cite{sundaresan2023learning,achiam2023gpt,liu2021swin}. Finally, we evaluate the real-world efficacy of our system for feeding a complete plate to a care recipient with mobility limitations.



\begin{figure*}[t]
    \includegraphics[width=\textwidth]{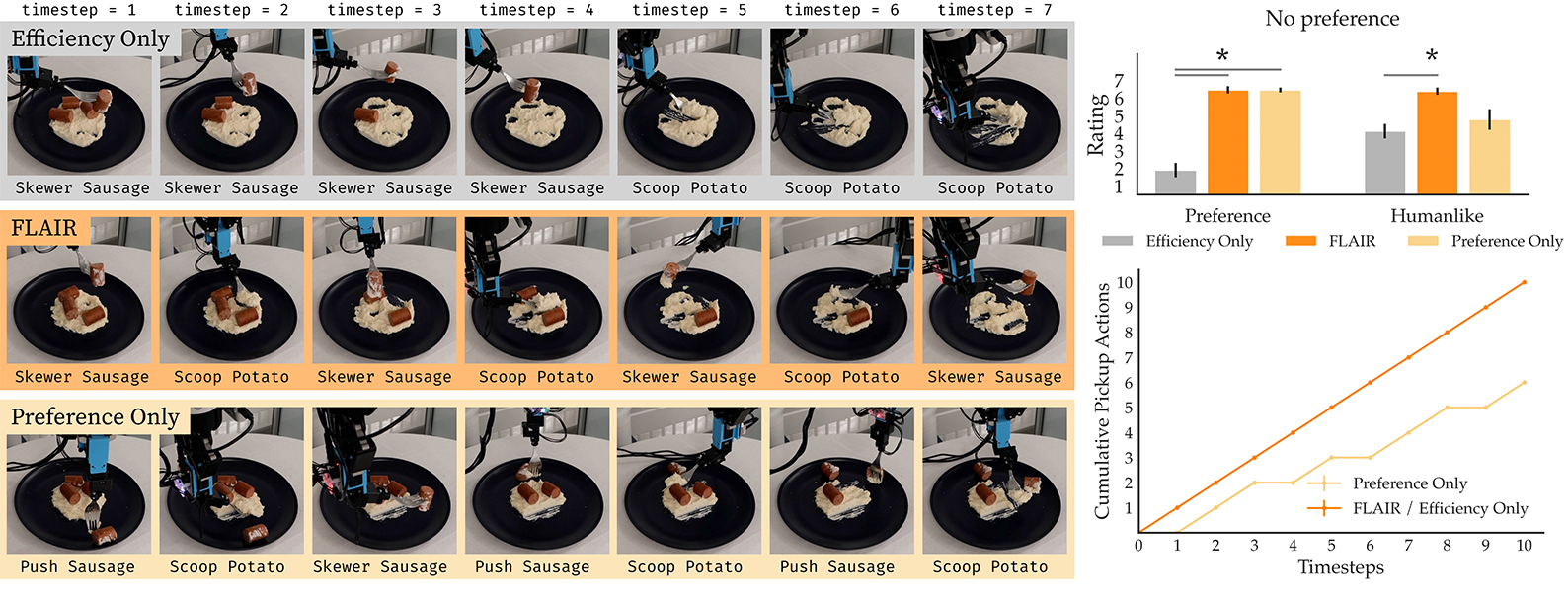}
    \vspace{-0.6cm}
    \caption{Example run on a plate with mashed potatoes and sausages where the user specified no preference. \flair, which balances user preferences (bite variety) and efficiency, is judged by users to better adhere to preferences than Efficiency-Only and outperforms Preference-Only (Commonsense-Only) in plate clearance. Consequently, \flair~is considered to provide a more human-like feeding experience compared to the baseline methods. Note that $^*$ indicates statistical significance ($p$-value $< 0.05$), determined via a Mann-Whitney U test.}
    \vspace{-0.5cm}
    \label{fig:example_run}
    \vspace{-0.2cm}
\end{figure*}

\subsection{Bite Acquisition Experiments}
\noindent \textbf{Baselines:} \flair~presents a unique approach of taking into account both preference and efficiency considerations for bite sequencing. This naturally begs the question of how an \emph{Efficiency-Only} or \emph{Preference-Only} approach would compare. We implement an Efficiency-Only baseline which greedily selects the next bite as the item which requires the least number of pre-acquisition and acquisition skills for pickup in the current instant, as dictated by the task planner $|\mathcal{T}(\cdot,\cdot)|$. The Preference-Only baseline is identical to \flair~in implementation, but notably omits efficiency scores when prompting the LLM to generate a next bite. This encourages the LLM to only respect a user's preference without consideration for how efficient a particular bite may be. In the case that a user has no preference for feeding, we refer to the Preference-Only baseline as \emph{Commonsense-Only.}

\smallskip
\noindent \textbf{Evaluation Plates:} We consider an evaluation suite of 6 diverse plates of food spanning a wide range of food categories, visualized in \cref{fig:plates}. We include 2 in-the-wild noodle dishes: a spaghetti and meatballs plate which is a prepared frozen meal from a grocery store, and a fettuccine alfredo dish with chicken and broccoli ordered from Applebee's via Doordash. We also consider 2 homemade semisolid dishes: mashed potatoes with sausage, and oatmeal with strawberries. Lastly, we evaluate an appetizer plate of strawberry, watermelon, celery, ranch, and chocolate dipping sauce, as well as a dessert plate of a whole banana, brownie bites, and chocolate dipping sauce.

\smallskip
\noindent \textbf{User Study Design:} We evaluate \flair's ability to cater to user preferences via a two-phase user study across 42 individuals without mobility limitations (Ages: 19-64, Genders: 22F, 20M). In the first phase, we present participants with a survey showing images of all 6 evaluation plates, and solicit their natural language preference over how they would prefer to be fed each plate. In the survey, we specify the capabilities of our skill library to the participants of our user study, and ask them to note preferences over their preferred order of bites, or pairings of food items with sauces. Details on the reported user preferences are provided in the Appendix. Since evaluating each submitted preference across all of the plates and baselines is not scalable, we cluster the submitted preferences into common shared responses via LLM summarization (GPT-4V). We focus on cases where users have either no preference or strong preferences, as slight preferences are not informative for comparing method behaviors. Thus, we specifically prompt GPT to filter for strong preferences (i.e. `Always feed me alternating bites of X and Y' or `Please do not feed me X') and group them accordingly. For each of the six plates, we then evaluate our system on the 2 most popular strong preferences summarized per plate, as well as a `I have no preference' setting for completeness.

We hypothesized the following:
\begin{itemize}
    \item \textbf{H1}: Compared to the Preference-Only baseline, \flair's consideration of efficiency in bite sequencing will lead to more number of bites across all settings. 
    \item \textbf{H2}: Compared to the Efficiency-Only baseline, \flair's consideration of user preferences in bite sequencing in presence of strong preferences will lead to more perceived adherence to preferences, and more human-like feeding, rated based on the statement ``This method is similar to the strategy I would use to feed myself.''
    \item \textbf{H3}: Compared to the Efficiency-Only baseline, \flair's consideration of commonsense reasoning in bite sequencing in the absence of preferences will lead to more perceived adherence to bite variety and common food item pairings, and more human-like feeding.
\end{itemize}
This user study was approved by the Institutional Review Boards of both Cornell University and Stanford University.

\begin{figure*}
    \centering
    \includegraphics[width=1.0\textwidth]{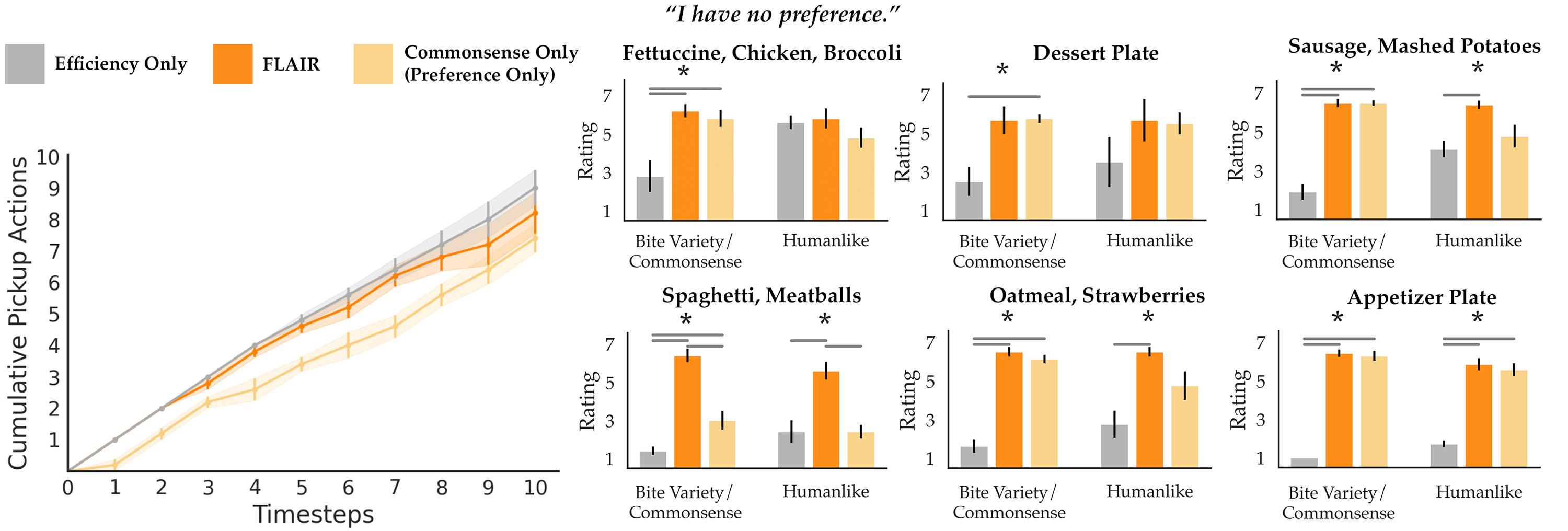}
    \vspace{-0.6cm}
    \caption{\textit{Left:} \flair picks up more bites than Preference-Only (Commonsense-Only) accumulated across all plates for no preference scenarios. \textit{Right:} For most no preference scenarios, average participant ratings show \flair~significantly outperforms the Efficiency-Only baseline in aligning with user preferences (bite variety) and achieving human-like feeding. Note that $^*$ indicates statistical significance ($p$-value $< 0.05$), determined via a Mann-Whitney U test.} 
    \vspace{-0.7cm}
    \label{fig:all_no_pref}
\end{figure*}

\smallskip
\noindent \textbf{Food Pickup Results:} \cref{fig:all_no_pref} displays the results of food pickup over time across methods in the no preference scenario. We provide additional results for food pickup efficiency for all methods averaged across all plates (both strong preferences and no preferences) in Appendix, noting a similar trend. Due to its consideration of efficiency in bite sequencing, \flair~executes a greater number of pickup skills compared to Preference-Only, validating \textbf{H1}. This is because when faced with multiple valid candidate bites, \flair, informed with efficiency scores for each bite, is able to choose the bite that optimizes for efficiency. In contrast, Preference-Only randomly selects one bite from this set, often leading to inefficient acquisition trajectories (\cref{fig:example_run}). The efficiency disparity between Efficiency-Only and \flair can be linked to settings where bite variety or strong preferences require the robot to pickup bites that are less efficient than those of a method which does not take such preferences into account. For instance, in a scenario where a robot is instructed to feed a bed of spaghetti hidden beneath multiple meatballs, methods that consider preferences must undertake multiple pre-acquisition skills to push away the meatballs. 

\begin{figure*}
    \centering
    \includegraphics[width=1.0\textwidth]{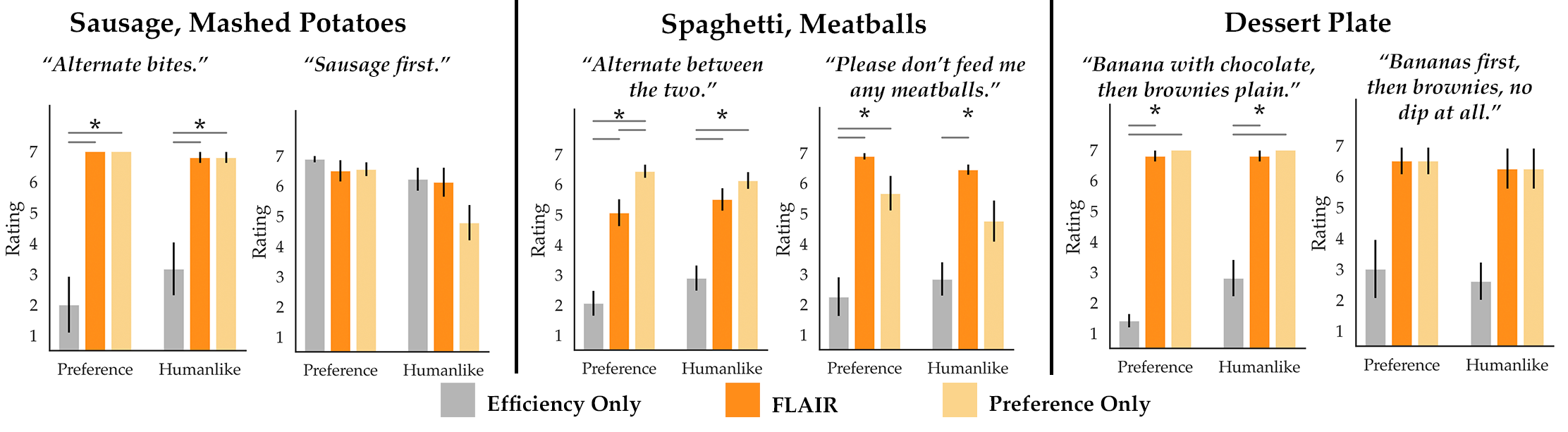}
    \vspace{-0.7cm}
    \caption{Average participant ratings for settings with strong user preferences show \flair~significantly outperforms Efficiency-Only baseline in aligning with user preferences and achieving human-like feeding in all scenarios, except cases where the efficiency-based bite sequence coincidentally aligns with user preferences. 
    Note that $^*$ indicates statistical significance ($p$-value $< 0.05$), determined via a Mann-Whitney U test.
    } 
    \vspace{-0.2cm}
    \label{fig:strong_prefs}
\end{figure*}

\smallskip
\noindent \textbf{User Evaluation:} \cref{fig:strong_prefs} presents average participant ratings comparing \flair~with baseline approaches for settings with strong user preferences. We conduct a Mann-Whitney U test for statistical significance, and indicate pairs of methods for which the average participant ratings were significant ($p$-value $< 0.05$). This is a non-parametric test compatible with ordinal Likert data, and without specific assumptions on the normality or variance of the data distributions. By integrating user preferences into task planning, \flair~substantially surpasses the Efficiency-Only baseline in terms of adherence to user preferences and human-like feeding across various settings, as hypothesized (\textbf{H2}). The exceptions, where the performance difference between Efficiency-Only and \flair is not significant, occur in settings where the bite sequence, generated based solely on efficiency, inadvertently matches the user's preferences. 
\cref{fig:all_no_pref} shows participant ratings that compare \flair with baseline approaches in settings where no user preferences were specified. By leveraging commonsense reasoning, \flair significantly outperforms the Efficiency-Only baseline across most plates by ensuring bite variety and appropriately pairing food items with dips, resulting in a more human-like feeding experience (\textbf{H3}).


\subsection{Comparisons with Task Planning Baselines}
For evaluating necessity of pre-acquisition actions, \flair first estimates the distribution of food items by sensing density and entropy metrics from segmented observations, and then uses a hierarchical decision-tree style approach. In this section, we compare this approach against other task planning baselines. The closest relevant work, VAPORS~\cite{sundaresan2023learning}, concentrates on noodle dishes and employs physics-based simulations for decision-making between twirling and grouping noodles. Our work, however, encompasses a broader spectrum of food textures and types (such as solids like meatballs, semi-solids like mashed potatoes, and noodle-like items like spaghetti), making direct adaptation of VAPORS challenging due to the complex physics simulations required for accurately representing their varied interactions. Taking this gap into account, we compare FLAIR’s task planning accuracy against 3 established baselines: (i) VAPORS, (ii) VLM-TaskPlanner, which queries a VLM (GPT-4V~\cite{achiam2023gpt}) using 10 in-context examples from the training set to decide between candidate actions, and (iii) image classification using a pre-trained Swin-Transformer~\cite{liu2021swin} fine-tuned on the training set. We leverage 2 datasets of plate images with next action ground truth labels as a test-bed:

\textbf{Evaluation on noodle-only plates from VAPORS (\cite{sundaresan2023learning}).} 
We use the $\sim100$ held-out noodle-only image dataset from VAPORS and have two third-party human annotators label each image with ``Twirl" or ``Group," corresponding to the action they deem appropriate. We extract the images with intercoder agreement and evenly split them into training and test sets. We compare our task planning approach (FLAIR), against the three baselines on this set. 

\begin{table}[b]
\centering
\vspace{-0.5cm}
\caption{Comparison of \flair's task planner with baselines}
\label{tab:classification_results}
\begin{tabular}{lcc}
    \hline
     & User Study Plates & Noodle-Only Plates \\ \hline
    \flair & \textbf{0.817} & \textbf{0.854} \\
    VAPORS~\cite{sundaresan2023learning} & - & 0.415 \\
    VLM-TaskPlanner~\cite{achiam2023gpt} & 0.518 & 0.683 \\ 
    Swin-Transformer \cite{liu2021swin} & 0.720 & 0.785 \\
    \hline
\end{tabular}
\end{table}

\textbf{Evaluation on logged user study plates.} We further compare FLAIR to baselines on the $\sim100$ images per plate logged from our user study. Each image is labeled by two third-party human annotators with appropriate ground truth labels for food items where pre-acquisition is relevant: twirl/group/push for noodles, push/scoop for semisolids (oatmeal/mashed potatoes), and cut/acquire for a whole banana. We extract the images with intercoder agreement, split them evenly into training and test sets, and report the mean accuracy across plates.

\flair uses identical parameters for all plates, whereas other baselines use plate-specific parameters inferred from their respective training data. \cref{tab:classification_results} shows the results. \flair significantly outperforms all other baselines on both datasets: the noodle-only and user study plates. We posit that VAPORS may suffer due to the sim-to-real gap present in real vs. simulated observations, and the black-box VLM-TaskPlanner struggles without reasoning in a hierarchical manner. The Swin-Transformer classifier is the most competitive baseline, but likely suffers due to a lack of large-scale training data.



\subsection{Demonstration of Real-World Feeding}

We demonstrate FLAIR's effectiveness in helping a care recipient with severe mobility restrictions eat an entree dish comprising boiled baby carrots, watermelon, strawberries, ranch dressing, and chocolate sauce. The care recipient, a 44-year-old Caucasian/White female with Multiple Sclerosis for 19 years, has a severely limited range of motion in their head and neck. Consequently, they require inside-mouth transfer~\cite{jenamani2024bitetransfer} of acquired bites for successful feeding.

\begin{figure}[t]
    \centering
    \includegraphics[width=0.43\textwidth]{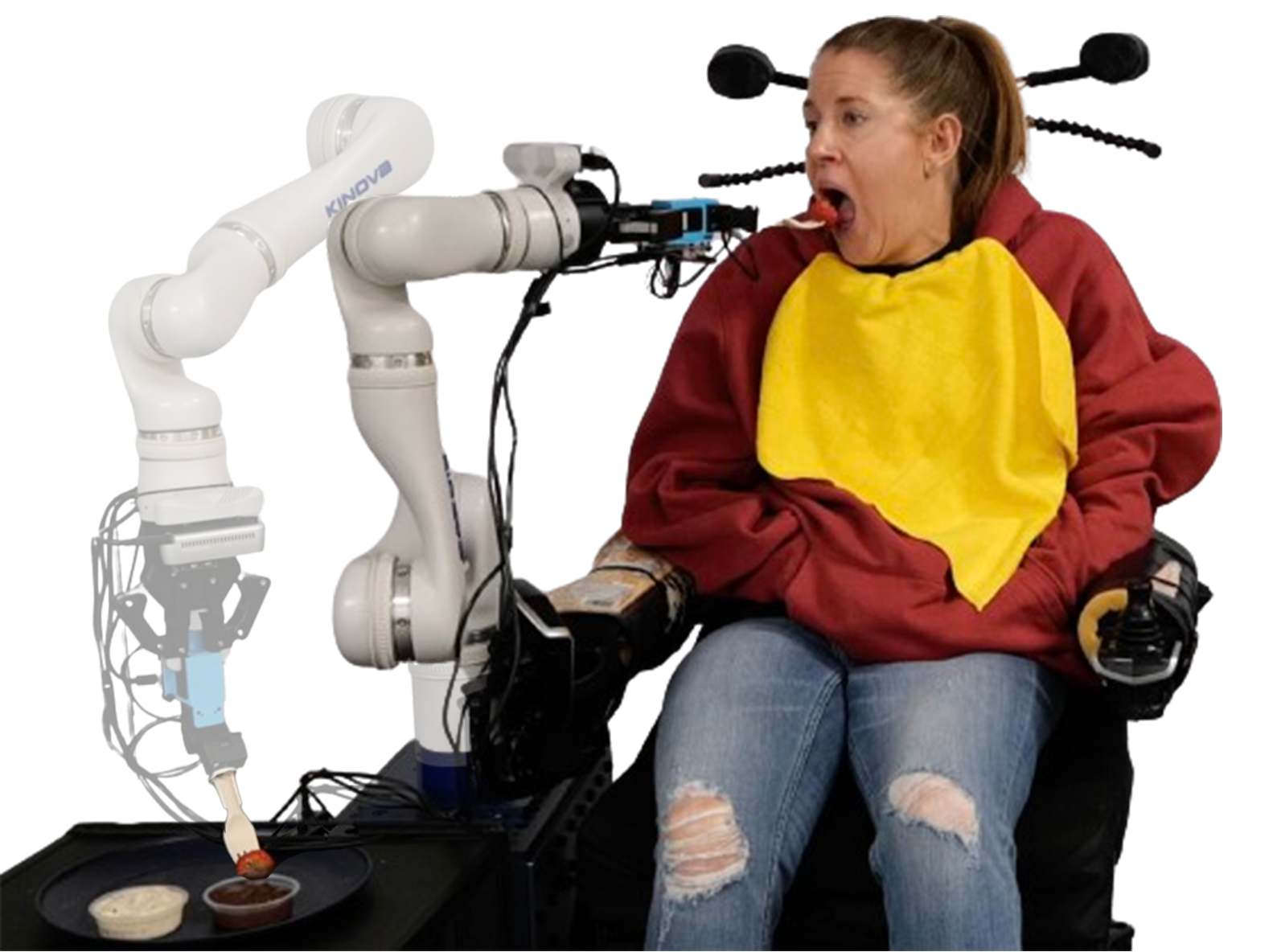}
    \caption{We demonstrate the real-world effectiveness of our method by feeding an entree dish to a care recipient with severe mobility limitations.} 
    \vspace{-20pt}
    \label{fig:kathleen}
\end{figure}

In the pre-study questionnaire, the care recipient mentioned that they typically have a preferred order in which they like to eat their meal. They convey this preference to their caregivers through natural language, and when caregivers adhere to this preference, it ``definitely enhances my eating experiences.'' For the considered plate, the care recipient specified ``I want to first finish all the celery with ranch dressing, then eat watermelon without any dips, and finally end with strawberries dipped in chocolate sauce.'' Adhering to this preference, FLAIR begins by skewering celery, dipping them in ranch dressing, and transferring them inside the mouth of the user. Once it detects there is no more celery on the plate, it switches to skewering watermelons and feeding them without dips as instructed. Finally, it skewers the strawberries, dips them in chocolate sauce, and feeds them to complete the meal (\cref{fig:kathleen}). Following successful feeding, we posed two questions on a seven-point Likert scale on the necessity of a robot-assisted feeding system to (i) have a diverse bite acquisition skill library, and (ii) adherence to meal preferences, for acceptance for day to day usage. The care recipient strongly agreed (rating = 7) with both, emphasizing the core contributions of our paper as critical aspects for an in-the-wild feeding system.

In the post-study questionnaire, the care recipient noted that while they often have specific preferences for the order in which they eat their food, they often refrain from sharing these preferences with their human caregivers. They expressed concern that such requests might impose additional burdens on caregivers who are already assisting them with feeding. However, they were hopeful that a robot designed to assist with feeding could accommodate their preferences seamlessly, thus enhancing their mealtime experience by respecting their autonomy and enabling them to better enjoy their meals.

\section{Discussion}

\flair is a first step towards robot-assisted feeding in real-world scenarios, adeptly handling various in-the-wild meals composed of diverse food items. We deploy \flair across 2 institutions and 3 different embodiments with a library of 7 dexterous skills. Our evaluations include both bite acquisition and bite transfer, along with a demonstration feeding a complete plate to a care recipient. \flair showcases the ability to abide by preferences across 42 individuals and a range of diverse plates, without compromising on efficient food pickup. Through our extensive evaluations, we identify the following limitations to guide future work in robot-assisted feeding. 

\smallskip
\noindent \textbf{Limitations of Food Perception using VLMs}. While current VLMs are capable of identifying food items on a plate, using these generated identifiers with open-set object detectors can sometimes lead to inaccuracies. FLAIR addresses this challenge by enriching the identifiers with a set of hand-coded descriptors tailored to the typical type of the food item, for example, specifying $\texttt{`banana'}$ as [$\texttt{`banana piece'}$, $\texttt{`sliced banana'}$] and $\texttt{`fettuccine'}$ as [$\texttt{`fettuccine pasta'}$, $\texttt{`fettuccine noodles'}$]. In the future, advancements in open-set object detection may eventually make such specific enhancements unnecessary.

\smallskip
\noindent \textbf{Limitations of Food Manipulation Skills.} 
FLAIR leverages a library of skills inspired by state-of-the-art food manipulation methods, but open challenges that occasionally occur include: slippage during skewering, failing to twirl noodles or scoop mashed potatoes into reasonable bite sizes, failing to cut tough items, and errors due to perception (erroneous depth sensing or imprecise food detection) which can cause manipulation imprecision. Although some of these failures can be addressed by re-trying (as long as the item is re-detected), these challenges can be mitigated in the future by making the skills themselves reactive, enabling adaptive utensil trajectories that adjust to food slippage or deformation on the fly.


\smallskip
\noindent \textbf{Limitations of Bite Sequencing.} 
We harness LLMs to plan efficient bite sequences that adhere to user preferences. However, today's language models can sometimes generate unrealistic or irrelevant outputs (``hallucinations"). In FLAIR, we reduce hallucinated artifacts in bite sequencing by using prompt-engineering strategies which we detail in the Appendix. However, even with templated prompts, FLAIR is still limited by the tendency of language models to occasionally neglect context, such as the manipulation efficiency of food items and their remaining portions. In the future, we are excited by structured prompting strategies ~\cite{dhuliawala2023chain} and incorporation of real-time corrections from the user~\cite{shi2024yell} to address these challenges.

\smallskip
Although these are current limitations, \textbf{FLAIR's modular system design allows for easy interchange of the perception/planning stacks or even skills themselves.} Thus, it will be able to take full advantage of future advances in VLMs or better low-level skill policies that are learned or engineered.


\section{ACKNOWLEDGEMENT}

This work was partly funded by NSF IIS \#2132846, CAREER \#2238792, and DARPA under Contract HR001120C0107. It was additionally supported by NSF awards \#2132847, \#2218760, the Office of Naval Research award \#N00014-21-1-2298, and AFOSR YIP. Priya Sundaresan is supported by an NSF GRFP and Maram Sakr is supported by the Natural Sciences and Engineering Research Council of Canada (NSERC). We would like to acknowledge Anthony Song, Pranav Thakkar, Eric Hu, and Karan Jha for their assistance with user studies and annotations for our task planning experiments.


\bibliographystyle{plainnat}
\bibliography{references/feeding, references/preferences, references/foundation_models}

\begin{thebibliography}{58}
\providecommand{\natexlab}[1]{#1}
\providecommand{\url}[1]{\texttt{#1}}
\expandafter\ifx\csname urlstyle\endcsname\relax
  \providecommand{\doi}[1]{doi: #1}\else
  \providecommand{\doi}{doi: \begingroup \urlstyle{rm}\Url}\fi

\bibitem[obi()]{obi}
{Meet Obi}.
\newblock \url{https://meetobi.com/}.
\newblock [Online; accessed 6-June-2022].

\bibitem[nea(2024)]{neater}
Neater eater robot, 2024.
\newblock URL \url{https://www.neater.co.uk/neater-eater-robotic}.
\newblock (Accessed: 1st January, 2024).

\bibitem[Achiam et~al.(2023)Achiam, Adler, Agarwal, Ahmad, Akkaya, Aleman, Almeida, Altenschmidt, Altman, Anadkat, et~al.]{achiam2023gpt}
Josh Achiam, Steven Adler, Sandhini Agarwal, Lama Ahmad, Ilge Akkaya, Florencia~Leoni Aleman, Diogo Almeida, Janko Altenschmidt, Sam Altman, Shyamal Anadkat, et~al.
\newblock Gpt-4 technical report.
\newblock \emph{arXiv preprint arXiv:2303.08774}, 2023.

\bibitem[Agia et~al.(2022)Agia, Migimatsu, Wu, and Bohg]{agia2022taps}
Christopher Agia, Toki Migimatsu, Jiajun Wu, and Jeannette Bohg.
\newblock Taps: Task-agnostic policy sequencing.
\newblock \emph{arXiv preprint arXiv:2210.12250}, 2022.

\bibitem[Belkhale et~al.(2022)Belkhale, Gordon, Chen, Srinivasa, Bhattacharjee, and Sadigh]{belkhale2022balancing}
Suneel Belkhale, Ethan~K Gordon, Yuxiao Chen, Siddhartha Srinivasa, Tapomayukh Bhattacharjee, and Dorsa Sadigh.
\newblock Balancing efficiency and comfort in robot-assisted bite transfer.
\newblock In \emph{2022 International Conference on Robotics and Automation (ICRA)}, pages 4757--4763. IEEE, 2022.

\bibitem[Bhattacharjee et~al.(2020)Bhattacharjee, Gordon, Scalise, Cabrera, Caspi, Cakmak, and Srinivasa]{bhattacharjee2020moreautonomy}
Tapomayukh Bhattacharjee, Ethan~K Gordon, Rosario Scalise, Maria~E Cabrera, Anat Caspi, Maya Cakmak, and Siddhartha~S Srinivasa.
\newblock Is more autonomy always better? exploring preferences of users with mobility impairments in robot-assisted feeding.
\newblock In \emph{2020 15th ACM/IEEE International Conference on Human-Robot Interaction (HRI)}, pages 181--190. IEEE, 2020.

\bibitem[Brohan et~al.(2022)Brohan, Brown, Carbajal, Chebotar, Dabis, Finn, Gopalakrishnan, Hausman, Herzog, Hsu, et~al.]{brohan2022rt}
Anthony Brohan, Noah Brown, Justice Carbajal, Yevgen Chebotar, Joseph Dabis, Chelsea Finn, Keerthana Gopalakrishnan, Karol Hausman, Alex Herzog, Jasmine Hsu, et~al.
\newblock Rt-1: Robotics transformer for real-world control at scale.
\newblock \emph{arXiv preprint arXiv:2212.06817}, 2022.

\bibitem[Brohan et~al.(2023)Brohan, Brown, Carbajal, Chebotar, Chen, Choromanski, Ding, Driess, Dubey, Finn, et~al.]{brohan2023rt}
Anthony Brohan, Noah Brown, Justice Carbajal, Yevgen Chebotar, Xi~Chen, Krzysztof Choromanski, Tianli Ding, Danny Driess, Avinava Dubey, Chelsea Finn, et~al.
\newblock Rt-2: Vision-language-action models transfer web knowledge to robotic control.
\newblock \emph{arXiv preprint arXiv:2307.15818}, 2023.

\bibitem[Brose et~al.(2010)Brose, Weber, Salatin, Grindle, Wang, Vazquez, and Cooper]{brose2010role}
Steven~W Brose, Douglas~J Weber, Ben~A Salatin, Garret~G Grindle, Hongwu Wang, Juan~J Vazquez, and Rory~A Cooper.
\newblock The role of assistive robotics in the lives of persons with disability.
\newblock \emph{American Journal of Physical Medicine \& Rehabilitation}, 89\penalty0 (6):\penalty0 509--521, 2010.

\bibitem[Brown et~al.(2020)Brown, Mann, Ryder, Subbiah, Kaplan, Dhariwal, Neelakantan, Shyam, Sastry, Askell, et~al.]{brown2020language}
Tom Brown, Benjamin Mann, Nick Ryder, Melanie Subbiah, Jared~D Kaplan, Prafulla Dhariwal, Arvind Neelakantan, Pranav Shyam, Girish Sastry, Amanda Askell, et~al.
\newblock Language models are few-shot learners.
\newblock \emph{Advances in neural information processing systems}, 33:\penalty0 1877--1901, 2020.

\bibitem[Canal et~al.(2019)Canal, Aleny{\`a}, and Torras]{canal2019adapting}
Gerard Canal, Guillem Aleny{\`a}, and Carme Torras.
\newblock Adapting robot task planning to user preferences: an assistive shoe dressing example.
\newblock \emph{Autonomous Robots}, 2019.

\bibitem[Canal et~al.(2021)Canal, Torras, and Aleny{\`a}]{canal2021preferences}
Gerard Canal, Carme Torras, and Guillem Aleny{\`a}.
\newblock Are preferences useful for better assistance? a physically assistive robotics user study.
\newblock \emph{THRI}, 2021.

\bibitem[Chen et~al.(2023)Chen, Arkin, Zhang, Roy, and Fan]{chen2023autotamp}
Yongchao Chen, Jacob Arkin, Yang Zhang, Nicholas Roy, and Chuchu Fan.
\newblock Autotamp: Autoregressive task and motion planning with llms as translators and checkers.
\newblock \emph{arXiv preprint arXiv:2306.06531}, 2023.

\bibitem[Chi{\`o} et~al.(2006)Chi{\`o}, Gauthier, Vignola, Calvo, Ghiglione, Cavallo, Terreni, and Mutani]{chio2006caregiver}
Adriano Chi{\`o}, A~Gauthier, A~Vignola, Andrea Calvo, Paolo Ghiglione, Enrico Cavallo, AA~Terreni, and Roberto Mutani.
\newblock Caregiver time use in als.
\newblock \emph{Neurology}, 67\penalty0 (5):\penalty0 902--904, 2006.

\bibitem[Chowdhery et~al.(2023)Chowdhery, Narang, Devlin, Bosma, Mishra, Roberts, Barham, Chung, Sutton, Gehrmann, et~al.]{chowdhery2023palm}
Aakanksha Chowdhery, Sharan Narang, Jacob Devlin, Maarten Bosma, Gaurav Mishra, Adam Roberts, Paul Barham, Hyung~Won Chung, Charles Sutton, Sebastian Gehrmann, et~al.
\newblock Palm: Scaling language modeling with pathways.
\newblock \emph{Journal of Machine Learning Research}, 24\penalty0 (240):\penalty0 1--113, 2023.

\bibitem[Devlin et~al.(2018)Devlin, Chang, Lee, and Toutanova]{devlin2018bert}
Jacob Devlin, Ming-Wei Chang, Kenton Lee, and Kristina Toutanova.
\newblock Bert: Pre-training of deep bidirectional transformers for language understanding.
\newblock \emph{arXiv preprint arXiv:1810.04805}, 2018.

\bibitem[Dhuliawala et~al.(2023)Dhuliawala, Komeili, Xu, Raileanu, Li, Celikyilmaz, and Weston]{dhuliawala2023chain}
Shehzaad Dhuliawala, Mojtaba Komeili, Jing Xu, Roberta Raileanu, Xian Li, Asli Celikyilmaz, and Jason Weston.
\newblock Chain-of-verification reduces hallucination in large language models.
\newblock \emph{arXiv preprint arXiv:2309.11495}, 2023.

\bibitem[Feng et~al.(2019)Feng, Kim, Lee, Gordon, Schmittle, Kumar, Bhattacharjee, and Srinivasa]{feng2019robot}
Ryan Feng, Youngsun Kim, Gilwoo Lee, Ethan~K Gordon, Matt Schmittle, Shivaum Kumar, Tapomayukh Bhattacharjee, and Siddhartha~S Srinivasa.
\newblock Robot-assisted feeding: Generalizing skewering strategies across food items on a plate.
\newblock In \emph{The International Symposium of Robotics Research}, pages 427--442. Springer, 2019.

\bibitem[Gallenberger et~al.(2019)Gallenberger, Bhattacharjee, Kim, and Srinivasa]{gallenberger2019transfer}
Daniel Gallenberger, Tapomayukh Bhattacharjee, Youngsun Kim, and Siddhartha~S Srinivasa.
\newblock Transfer depends on acquisition: Analyzing manipulation strategies for robotic feeding.
\newblock In \emph{2019 14th ACM/IEEE International Conference on Human-Robot Interaction (HRI)}, pages 267--276. IEEE, 2019.

\bibitem[Gao et~al.(2023)Gao, Sarkar, Xia, Xiao, Wu, Ichter, Majumdar, and Sadigh]{gao2023physically}
Jensen Gao, Bidipta Sarkar, Fei Xia, Ted Xiao, Jiajun Wu, Brian Ichter, Anirudha Majumdar, and Dorsa Sadigh.
\newblock Physically grounded vision-language models for robotic manipulation.
\newblock \emph{arXiv preprint arXiv:2309.02561}, 2023.

\bibitem[Gordon et~al.(2020)Gordon, Meng, Bhattacharjee, Barnes, and Srinivasa]{gordon2020adaptive}
Ethan~K Gordon, Xiang Meng, Tapomayukh Bhattacharjee, Matt Barnes, and Siddhartha~S Srinivasa.
\newblock Adaptive robot-assisted feeding: An online learning framework for acquiring previously unseen food items.
\newblock In \emph{2020 IEEE/RSJ International Conference on Intelligent Robots and Systems (IROS)}, pages 9659--9666. IEEE, 2020.

\bibitem[Gordon et~al.(2021)Gordon, Roychowdhury, Bhattacharjee, Jamieson, and Srinivasa]{gordon2021leveraging}
Ethan~K Gordon, Sumegh Roychowdhury, Tapomayukh Bhattacharjee, Kevin Jamieson, and Siddhartha~S Srinivasa.
\newblock Leveraging post hoc context for faster learning in bandit settings with applications in robot-assisted feeding.
\newblock In \emph{2021 IEEE International Conference on Robotics and Automation (ICRA)}, pages 10528--10535. IEEE, 2021.

\bibitem[Gordon et~al.(2023)Gordon, Nanavati, Challa, Zhu, Faulkner, and Srinivasa]{gordon2023towards}
Ethan~Kroll Gordon, Amal Nanavati, Ramya Challa, Bernie~Hao Zhu, Taylor Annette~Kessler Faulkner, and Siddhartha Srinivasa.
\newblock Towards general single-utensil food acquisition with human-informed actions.
\newblock In \emph{Conference on Robot Learning}, pages 2414--2428. PMLR, 2023.

\bibitem[Grannen et~al.(2022)Grannen, Wu, Belkhale, and Sadigh]{grannen2022learning}
Jennifer Grannen, Yilin Wu, Suneel Belkhale, and Dorsa Sadigh.
\newblock Learning bimanual scooping policies for food acquisition.
\newblock In \emph{6th Annual Conference on Robot Learning}, 2022.
\newblock URL \url{https://openreview.net/forum?id=qDtbMK67PJG}.

\bibitem[Huang et~al.(2022)Huang, Xia, Xiao, Chan, Liang, Florence, Zeng, Tompson, Mordatch, Chebotar, et~al.]{huang2022inner}
Wenlong Huang, Fei Xia, Ted Xiao, Harris Chan, Jacky Liang, Pete Florence, Andy Zeng, Jonathan Tompson, Igor Mordatch, Yevgen Chebotar, et~al.
\newblock Inner monologue: Embodied reasoning through planning with language models.
\newblock \emph{arXiv preprint arXiv:2207.05608}, 2022.

\bibitem[Huang et~al.(2023)Huang, Wang, Zhang, Li, Wu, and Fei-Fei]{huang2023voxposer}
Wenlong Huang, Chen Wang, Ruohan Zhang, Yunzhu Li, Jiajun Wu, and Li~Fei-Fei.
\newblock Voxposer: Composable 3d value maps for robotic manipulation with language models.
\newblock \emph{arXiv preprint arXiv:2307.05973}, 2023.

\bibitem[Jacobsson et~al.(2000)Jacobsson, Axelsson, Österlind, and Norberg]{jacobsson2000eatingprocess}
Catrine Jacobsson, Karin Axelsson, Per~Olov Österlind, and Astrid Norberg.
\newblock How people with stroke and healthy older people experience the eating process.
\newblock \emph{Journal of Clinical Nursing}, 9\penalty0 (2):\penalty0 255--264, 2000.
\newblock \doi{https://doi.org/10.1046/j.1365-2702.2000.00355.x}.
\newblock URL \url{https://onlinelibrary.wiley.com/doi/abs/10.1046/j.1365-2702.2000.00355.x}.

\bibitem[Jenamani et~al.(2024)Jenamani, Stabile, Liu, Anwar, Dimitropoulou, and Bhattacharjee]{jenamani2024bitetransfer}
Rajat~Kumar Jenamani, Daniel Stabile, Ziang Liu, Abrar Anwar, Katherine Dimitropoulou, and Tapomayukh Bhattacharjee.
\newblock Feel the bite: Robot-assisted inside-mouth bite transfer using robust mouth perception and physical interaction-aware control.
\newblock In \emph{Proceedings of the 2024 ACM/IEEE International Conference on Human-Robot Interaction}, pages 313--322, 2024.

\bibitem[Jiang et~al.(2024)Jiang, Huang, Wu, Li, Garg, Nayyeri, Wang, and Li]{jiang2024roboexp}
Hanxiao Jiang, Binghao Huang, Ruihai Wu, Zhuoran Li, Shubham Garg, Hooshang Nayyeri, Shenlong Wang, and Yunzhu Li.
\newblock Roboexp: Action-conditioned scene graph via interactive exploration for robotic manipulation.
\newblock \emph{arXiv preprint arXiv:2402.15487}, 2024.

\bibitem[Kambhampati(2024)]{kambhampati2024can}
Subbarao Kambhampati.
\newblock Can large language models reason and plan?
\newblock \emph{Annals of the New York Academy of Sciences}, 2024.

\bibitem[Karamcheti et~al.(2023)Karamcheti, Nair, Chen, Kollar, Finn, Sadigh, and Liang]{karamcheti2023language}
Siddharth Karamcheti, Suraj Nair, Annie~S Chen, Thomas Kollar, Chelsea Finn, Dorsa Sadigh, and Percy Liang.
\newblock Language-driven representation learning for robotics.
\newblock \emph{arXiv preprint arXiv:2302.12766}, 2023.

\bibitem[Keely et~al.(2024)Keely, Nemlekar, and Losey]{keely2024kiri}
Maya~N Keely, Heramb Nemlekar, and Dylan~P Losey.
\newblock Kiri-spoon: A soft shape-changing utensil for robot-assisted feeding.
\newblock \emph{arXiv preprint arXiv:2403.05784}, 2024.

\bibitem[Kirillov et~al.(2023)Kirillov, Mintun, Ravi, Mao, Rolland, Gustafson, Xiao, Whitehead, Berg, Lo, et~al.]{kirillov2023segment}
Alexander Kirillov, Eric Mintun, Nikhila Ravi, Hanzi Mao, Chloe Rolland, Laura Gustafson, Tete Xiao, Spencer Whitehead, Alexander~C Berg, Wan-Yen Lo, et~al.
\newblock Segment anything.
\newblock In \emph{Proceedings of the IEEE/CVF International Conference on Computer Vision}, pages 4015--4026, 2023.

\bibitem[Kwon et~al.(2023)Kwon, Hu, Myers, Karamcheti, Dragan, and Sadigh]{kwon2023toward}
Minae Kwon, Hengyuan Hu, Vivek Myers, Siddharth Karamcheti, Anca Dragan, and Dorsa Sadigh.
\newblock Toward grounded social reasoning.
\newblock \emph{arXiv preprint arXiv:2306.08651}, 2023.

\bibitem[Li et~al.(2022)Li, Li, Xiong, and Hoi]{li2022blip}
Junnan Li, Dongxu Li, Caiming Xiong, and Steven Hoi.
\newblock Blip: Bootstrapping language-image pre-training for unified vision-language understanding and generation.
\newblock In \emph{International conference on machine learning}, pages 12888--12900. PMLR, 2022.

\bibitem[Liang et~al.(2023)Liang, Huang, Xia, Xu, Hausman, Ichter, Florence, and Zeng]{liang2023code}
Jacky Liang, Wenlong Huang, Fei Xia, Peng Xu, Karol Hausman, Brian Ichter, Pete Florence, and Andy Zeng.
\newblock Code as policies: Language model programs for embodied control.
\newblock In \emph{2023 IEEE International Conference on Robotics and Automation (ICRA)}, pages 9493--9500. IEEE, 2023.

\bibitem[Liu et~al.(2024)Liu, Fang, Abbeel, and Levine]{liu2024moka}
Fangchen Liu, Kuan Fang, Pieter Abbeel, and Sergey Levine.
\newblock Moka: Open-vocabulary robotic manipulation through mark-based visual prompting.
\newblock \emph{arXiv preprint arXiv:2403.03174}, 2024.

\bibitem[Liu et~al.(2023)Liu, Zeng, Ren, Li, Zhang, Yang, Li, Yang, Su, Zhu, et~al.]{liu2023grounding}
Shilong Liu, Zhaoyang Zeng, Tianhe Ren, Feng Li, Hao Zhang, Jie Yang, Chunyuan Li, Jianwei Yang, Hang Su, Jun Zhu, et~al.
\newblock Grounding dino: Marrying dino with grounded pre-training for open-set object detection.
\newblock \emph{arXiv preprint arXiv:2303.05499}, 2023.

\bibitem[Liu et~al.(2021)Liu, Lin, Cao, Hu, Wei, Zhang, Lin, and Guo]{liu2021swin}
Ze~Liu, Yutong Lin, Yue Cao, Han Hu, Yixuan Wei, Zheng Zhang, Stephen Lin, and Baining Guo.
\newblock Swin transformer: Hierarchical vision transformer using shifted windows.
\newblock In \emph{Proceedings of the IEEE/CVF international conference on computer vision}, pages 10012--10022, 2021.

\bibitem[Madan et~al.(2022)Madan, Jenamani, Nguyen, Moustafa, Hu, Dimitropoulou, and Bhattacharjee]{madan2022sparcs}
Rishabh Madan, Rajat~Kumar Jenamani, Vy~Thuy Nguyen, Ahmed Moustafa, Xuefeng Hu, Katherine Dimitropoulou, and Tapomayukh Bhattacharjee.
\newblock Sparcs: Structuring physically assistive robotics for caregiving with stakeholders-in-the-loop.
\newblock In \emph{2022 IEEE/RSJ International Conference on Intelligent Robots and Systems (IROS)}, pages 641--648. IEEE, 2022.

\bibitem[Nanavati et~al.(2023)Nanavati, Alves-Oliveira, Schrenk, Gordon, Cakmak, and Srinivasa]{nanavati2023design}
Amal Nanavati, Patricia Alves-Oliveira, Tyler Schrenk, Ethan~K Gordon, Maya Cakmak, and Siddhartha~S Srinivasa.
\newblock Design principles for robot-assisted feeding in social contexts.
\newblock In \emph{Proceedings of the 2023 ACM/IEEE International Conference on Human-Robot Interaction}, pages 24--33, 2023.

\bibitem[Padalkar et~al.(2023)Padalkar, Pooley, Jain, Bewley, Herzog, Irpan, Khazatsky, Rai, Singh, Brohan, et~al.]{padalkar2023open}
Abhishek Padalkar, Acorn Pooley, Ajinkya Jain, Alex Bewley, Alex Herzog, Alex Irpan, Alexander Khazatsky, Anant Rai, Anikait Singh, Anthony Brohan, et~al.
\newblock Open x-embodiment: Robotic learning datasets and rt-x models.
\newblock \emph{arXiv preprint arXiv:2310.08864}, 2023.

\bibitem[Park et~al.(2020)Park, Hoshi, Mahajan, Kim, Erickson, Rogers, and Kemp]{park2020active}
Daehyung Park, Yuuna Hoshi, Harshal~P Mahajan, Ho~Keun Kim, Zackory Erickson, Wendy~A Rogers, and Charles~C Kemp.
\newblock Active robot-assisted feeding with a general-purpose mobile manipulator: Design, evaluation, and lessons learned.
\newblock \emph{Robotics and Autonomous Systems}, 124:\penalty0 103344, 2020.

\bibitem[Radford et~al.(2018)Radford, Narasimhan, Salimans, Sutskever, et~al.]{radford2018improving}
Alec Radford, Karthik Narasimhan, Tim Salimans, Ilya Sutskever, et~al.
\newblock Improving language understanding by generative pre-training.
\newblock 2018.

\bibitem[Radford et~al.(2019)Radford, Wu, Child, Luan, Amodei, Sutskever, et~al.]{radford2019language}
Alec Radford, Jeffrey Wu, Rewon Child, David Luan, Dario Amodei, Ilya Sutskever, et~al.
\newblock Language models are unsupervised multitask learners.
\newblock \emph{OpenAI blog}, 1\penalty0 (8):\penalty0 9, 2019.

\bibitem[Radford et~al.(2021)Radford, Kim, Hallacy, Ramesh, Goh, Agarwal, Sastry, Askell, Mishkin, Clark, et~al.]{radford2021learning}
Alec Radford, Jong~Wook Kim, Chris Hallacy, Aditya Ramesh, Gabriel Goh, Sandhini Agarwal, Girish Sastry, Amanda Askell, Pamela Mishkin, Jack Clark, et~al.
\newblock Learning transferable visual models from natural language supervision.
\newblock In \emph{International conference on machine learning}, pages 8748--8763. PMLR, 2021.

\bibitem[Sap et~al.(2020)Sap, Shwartz, Bosselut, Choi, and Roth]{sap2020commonsense}
Maarten Sap, Vered Shwartz, Antoine Bosselut, Yejin Choi, and Dan Roth.
\newblock Commonsense reasoning for natural language processing.
\newblock In \emph{Proceedings of the 58th Annual Meeting of the Association for Computational Linguistics: Tutorial Abstracts}, pages 27--33, 2020.

\bibitem[Shaikewitz et~al.(2022)Shaikewitz, Wu, Belkhale, Grannen, Sundaresan, and Sadigh]{shaikewitz2022mouth}
Lorenzo Shaikewitz, Yilin Wu, Suneel Belkhale, Jennifer Grannen, Priya Sundaresan, and Dorsa Sadigh.
\newblock In-mouth robotic bite transfer with visual and haptic sensing.
\newblock \emph{arXiv preprint arXiv:2211.12705}, 2022.

\bibitem[Shi et~al.(2024)Shi, Hu, Zhao, Sharma, Pertsch, Luo, Levine, and Finn]{shi2024yell}
Lucy~Xiaoyang Shi, Zheyuan Hu, Tony~Z Zhao, Archit Sharma, Karl Pertsch, Jianlan Luo, Sergey Levine, and Chelsea Finn.
\newblock Yell at your robot: Improving on-the-fly from language corrections.
\newblock \emph{arXiv preprint arXiv:2403.12910}, 2024.

\bibitem[Shune(2020)]{shune2020experience}
Samantha~E. Shune.
\newblock An altered eating experience: Attitudes toward feeding assistance among younger and older adults.
\newblock \emph{Rehabilitation nursing : the official journal of the Association of Rehabilitation Nurses}, 2020.

\bibitem[Sundaresan et~al.(2022)Sundaresan, Belkhale, and Sadigh]{sundaresan2022learning}
Priya Sundaresan, Suneel Belkhale, and Dorsa Sadigh.
\newblock Learning visuo-haptic skewering strategies for robot-assisted feeding.
\newblock In \emph{6th Annual Conference on Robot Learning}, 2022.
\newblock URL \url{https://openreview.net/forum?id=lLq09gVoaTE}.

\bibitem[Sundaresan et~al.(2023)Sundaresan, Wu, and Sadigh]{sundaresan2023learning}
Priya Sundaresan, Jiajun Wu, and Dorsa Sadigh.
\newblock Learning sequential acquisition policies for robot-assisted feeding.
\newblock In \emph{Conference on Robot Learning}, pages 1282--1299. PMLR, 2023.

\bibitem[Tai et~al.(2023)Tai, Chiu, Chao, and Chen]{tai2023scone}
Yen-Ling Tai, Yu~Chien Chiu, Yu-Wei Chao, and Yi-Ting Chen.
\newblock Scone: A food scooping robot learning framework with active perception.
\newblock In \emph{Conference on Robot Learning}, pages 849--865. PMLR, 2023.

\bibitem[Taylor(2018)]{taylor2018americans}
Danielle~M Taylor.
\newblock Americans with disabilities: 2014.
\newblock \emph{US Census Bureau}, pages 1--32, 2018.

\bibitem[Wu et~al.(2023)Wu, Antonova, Kan, Lepert, Zeng, Song, Bohg, Rusinkiewicz, and Funkhouser]{wu2023tidybot}
Jimmy Wu, Rika Antonova, Adam Kan, Marion Lepert, Andy Zeng, Shuran Song, Jeannette Bohg, Szymon Rusinkiewicz, and Thomas Funkhouser.
\newblock Tidybot: Personalized robot assistance with large language models.
\newblock \emph{arXiv preprint arXiv:2305.05658}, 2023.

\bibitem[Yang et~al.(2021)Yang, Wang, Yang, and Shi]{yang2021desire}
Guang Yang, Shuoyu Wang, Junyou Yang, and Peng Shi.
\newblock Desire-driven reasoning considering personalized care preferences.
\newblock \emph{Transactions on Systems, Man, and Cybernetics: Systems}, 2021.

\bibitem[Zha et~al.(2023)Zha, Cui, Lin, Kwon, Arenas, Zeng, Xia, and Sadigh]{zha2023distilling}
Lihan Zha, Yuchen Cui, Li-Heng Lin, Minae Kwon, Montserrat~Gonzalez Arenas, Andy Zeng, Fei Xia, and Dorsa Sadigh.
\newblock Distilling and retrieving generalizable knowledge for robot manipulation via language corrections.
\newblock \emph{arXiv preprint arXiv:2311.10678}, 2023.

\bibitem[Zhang et~al.(2019)Zhang, Sharma, Veloso, and Kroemer]{zhang2019leveraging}
Kevin Zhang, Mohit Sharma, Manuela Veloso, and Oliver Kroemer.
\newblock Leveraging multimodal haptic sensory data for robust cutting.
\newblock In \emph{2019 IEEE-RAS 19th International Conference on Humanoid Robots (Humanoids)}, 2019.

\end{thebibliography}

\newpage

\twocolumn[
\centering
\Huge{Appendix} \\
\Large{FLAIR: Feeding via Long-horizon AcquIsition of Realistic dishes} \\
\href{http://emprise.cs.cornell.edu/flair}{http://emprise.cs.cornell.edu/flair} 
\vspace{1em} 
]

\setcounter{section}{0} 
\renewcommand\thesection{\Alph{section}}
\renewcommand\thesubsection{\Alph{section}.\arabic{subsection}} 

\section{Hardware System Details}

The experiments described in this work have been conducted at two different universities, utilizing three different robot arms, each equipped with a specialized fork attachment and sensors. The design for the fork attachment, employed by both institutions, is adapted from Shaikewitz et al.~\cite{shaikewitz2022mouth} to accommodate the Robotiq 2F-85 gripper and to facilitate effective cutting actions. This fork attachment is illustrated in the main text of this paper and is publicly available as open-source hardware on the website. Specifically, the system at Stanford University employs both a Kinova Gen3 7-DoF arm and a Franka Emika Panda 7-DoF arm, both mounted directly onto tables (\cref{fig:hardware_setup}). In contrast, the setup at Cornell University utilizes a Kinova Gen3 6-DoF robotic arm, which is mounted on a Vention stand adjacent to a feeding table.

Both institutions perform experiments to showcase how our bite acquisition framework can be integrated with an outside-mouth bite transfer method~\cite{gallenberger2019transfer} with visual servoing (\cref{fig:servo}). Additionally, we use the setup at Cornell University to feed a care recipient with Multiple Sclerosis, demonstrating integration with an inside-mouth bite transfer method~\cite{jenamani2024bitetransfer}, which is necessary for feeding people with severe mobility limitations who cannot lean forward to take a bite. The successful execution of acquisition and transfer trials at both institutions, utilizing different robotic setups, underscores the reproducibility of our methodology and hardware.

\section{Prompting Details}

\subsection{Food Detection}
We use GPT-4V~\cite{achiam2023gpt} to generate a set of candidate labels for open-set object detectors, using the following prompt with accompanying images shown in \cref{fig:object_recognition}:

\begin{figure}[b]
\vspace{-0.6cm}
  \centering
  \begin{minipage}[b]{0.14\textwidth}
    \includegraphics[width=\textwidth]{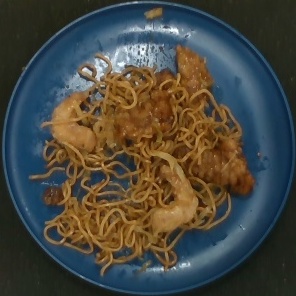}
  \end{minipage}
  \hfill 
  \begin{minipage}[b]{0.14\textwidth}
    \includegraphics[width=\textwidth]{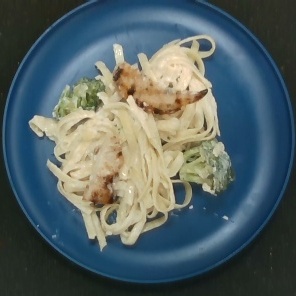}
  \end{minipage}
  \hfill 
  \begin{minipage}[b]{0.14\textwidth}
    \includegraphics[width=\textwidth]{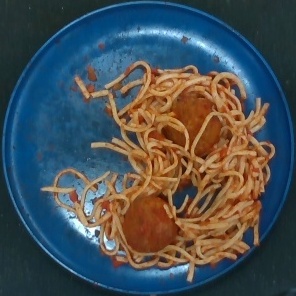}
  \end{minipage}
  \caption{Prompt images used for food item recognition with a VLM.}
  \vspace{-0.27cm}
  \label{fig:object_recognition}
\end{figure}

\vspace{-0.1cm}
\begin{tcolorbox}[colback=orange!10, colframe=gray!40, boxrule=00pt, sharp corners, fontupper=\small, left=0pt,]
\begin{verbatim}
Objective: To determine the food items on the plate.
Available items: ['shrimp', 'chicken', 'noodles', 
'chicken', 'broccoli', 'fettucine', 'meatball', 
'spaghetti', 'brownie', 'chocolate sauce', 'strawberry', 
'celery', 'watermelon', 'ranch dressing', 
'mashed potatoes', 'sausage']
The first 3 images are in-context examples with the ground 
truth food items.
Example Image 1: ['shrimp', 'chicken', 'noodles']
Example Image 2: ['chicken', 'broccoli', 'fettucine']
Example Image 3: ['meatball', 'spaghetti']
\end{verbatim}
\end{tcolorbox}

\begin{tcolorbox}[colback=orange!10, colframe=gray!40, boxrule=00pt, sharp corners, fontupper=\small, left=0pt,]
\begin{verbatim}
The 4th image is the test image. From the available items, 
what food items are present, as a comma-separated list? 
Format your answer in one-line as follows: ['first_item', 
'second_item', ..., 'last_item']
\end{verbatim}
\end{tcolorbox}

\subsection{Bite Sequencing}

We use the following prompt for bite sequencing, instructing the language model to balance user preferences, commonsense orderings, and efficiency. For Preference-Only, we omit efficiency from the prompt. While LLMs offer powerful commonsense and few-shot reasoning, they sometimes produce unrealistic outputs or overlook critical context, such as manipulation efficiency and remaining portions of each food item. To guide the LLM's reasoning, we use in-context examples and meticulously identify and analyze potential failure modes through extensive pre-study testing. We update our prompt to explicitly instruct the LLM to reduce these errors (underlined).

\begin{figure*}[t]
    \centering
    \makebox[\textwidth][c]{%
    \includegraphics[width=1\textwidth]{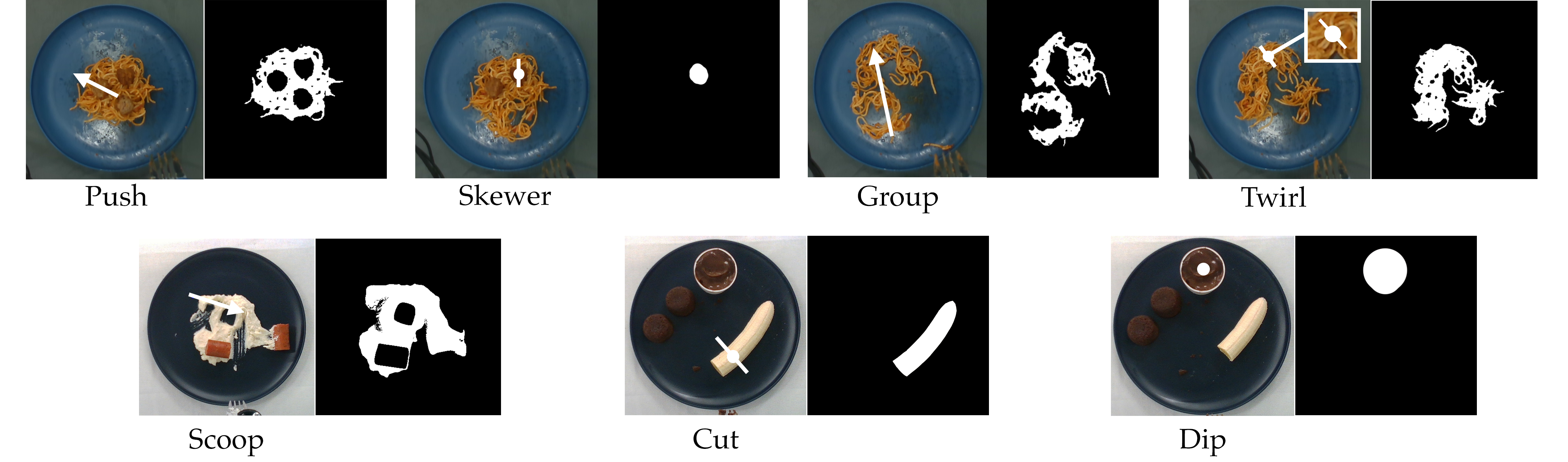}
    }
    \vspace{-19pt}
    \caption{\textbf{Skill Parameterizations}: We visualize the parameterization of each skill in \flair's skill library. White arrows indicate the direction of pushing, grouping, or scooping, while a white dot with a line through it represents the intended fork position and orientation, respectively.}
    \label{fig:parameterization}
    \vspace{-16pt}
\end{figure*}

\vspace{-0.1cm}
\begin{tcolorbox}[colback=orange!10, colframe=gray!40, boxrule=00pt, sharp corners, fontupper=\small, left=0pt,]
\begin{Verbatim}[commandchars=+\{\}]
You are feeding me a plate of food.
Items remaining: %s
Portions remaining (corresponding to items remaining 
list): %s 
Efficiency: %s
Preference: %s
Dipping sauces remaining: %s
---
Given this information, you will decide what bite to 
feed me next. You may ONLY suggest a bite from the 
available 'Items remaining', optionally dipped in an 
item from 'Dipping sauces.'
1) First, summarize what food items are remaining on 
the  plate with corresponding portions. +underline{Make sure that} 
+underline{your  summarization matches with the 'Items remaining'} 
+underline{and 'Portions remaining' provided at the beginning.}
2) Next, summarize your high-level strategy for feeding 
me. Describe the order of bites you would use to feed 
me the remaining items, and explain how you make your 
decision. +underline{Note that bites taken so far do not affect}
+underline{items remaining}. 
Decision Criteria:
- Prioritize my preference above all else. If I have a 
strong preference for a particular ordering of foods, 
you should obey that as best as possible. Respect user 
preferences for not dipping certain food items in 
certain sauces. If the preference specifies 'Feed me 
X and then Y', perceive it as me asking you to feed 
all of X before you feed any of Y.
- "Alternating between X and Y" means you can start 
with either X or Y, unless otherwise specified. If one 
item is more efficient, start with that.
- In the past, I have eaten the following bites: %s.
The first item of this list is the first bite I had,
the second item is the second bite, and so on. The last
element is the most recent bite that I just had. Use 
this to inform what you feed me next. +underline{Note that bites }
\end{Verbatim}
\end{tcolorbox}

\begin{tcolorbox}[colback=orange!10, colframe=gray!40, boxrule=00pt, sharp corners, fontupper=\small, left=0pt,]
\begin{Verbatim}[commandchars=+\{\}] 
+underline{taken so far are past bites and thus do not affect the }
+underline{specified items or portions remaining on the plate.}
- In your answer, state the portions of each item 
remaining. +underline{Ensure that this matches with the 'Portions} 
+underline{remaining' provided at the beginning. Note that these} 
+underline{items haven't been consumed.} If I do not have a strong 
preference, use commonsense. For dishes like noodles 
with vegetables and meat, or porridge with different 
toppings, prioritize feeding me a MIX of different 
items. This means do not feed me the same item 
consecutively twice if it doesn't affect efficiency. 
The exception to this rule is if there is a lot more of 
one item than another. In that case, you should 
prioritize feeding me the item with more quantity, 
until the portions of items left are more evenly 
distributed, even if I have eaten the larger portion 
item consecutively. If only a single portion of an item 
is left, do not feed it but leave it until later for 
better bite variety (unless the preference instructs 
otherwise). Lastly, use your best judgment to figure 
out what ordering makes sense. For instance, looking 
at a plate of carrots, ranch, cantaloupe, apples, and
caramel, you should be able to tell that apples go 
with the caramel, and carrots go with the ranch, 
and that apple should not be dipped in ranch because 
that is not typical, and cantaloupe and other melon 
types should be eaten plain. Avoid feeding savory 
foods with sweet sauces, and vice versa (i.e. chicken 
nugget and chocolate sauce is NOT a good suggestion, 
apple should not be dipped in ketchup; cantaloupe, 
honeydew, and melons are usually eaten plain without 
sauce). A good rule of thumb is to prioritize: common 
pairings (i.e. sweet savory), portion sizes, and bite 
variability (in that order).
3) Next, choose an item from 'Items available' to feed 
next, based on your answer to 2). Examples: "Feed 
shrimp", "Feed apple dipped in caramel", "Do not feed a 
bite". You can either feed a single food item which 
appears in 'Items remaining' ("Feed shrimp"), a single 
food item dipped in something ("Feed apple dipped in 
caramel"), or opt to not feed an item ("Do not feed a 
bite"); only do this if I strongly dislike the 
remaining bites, or I only requested a 2 bites of 
'tiramisu,' for example, at the end of a meal, and you
already fed me both (as mentioned in 'The bites I have 
taken so far'). +underline{You CANNOT feed a dip (i.e. ranch,} 
+underline{mustard, whipped cream, bbq sauce, chocolate sauce,} 
+underline{etc.) by itself. You can ONLY feed an item if it} 
\end{Verbatim}
\end{tcolorbox}

\begin{tcolorbox}[colback=orange!10, colframe=gray!40, boxrule=00pt, sharp corners, fontupper=\small, left=0pt,]
\begin{Verbatim}[commandchars=+\{\}]
+underline{is present in 'Items remaining.' For example, given}
+underline{'Items remaining: ["broccoli", "asparagus"]', you} 
+underline{cannot suggest "Feed pasta", even if I have taken a} 
+underline{bite of 'pasta' in the past.}
4) Now, consider efficiency; in the numbers listed 
above, +underline{higher means that food item is more efficient to} 
+underline{pick up}. Would you change your answer to 3) given this 
information? If the portions are even (difference of <3 
bites), and my preference is weak or alternating, and 
one item is clearly more efficient, you may change your
answer. Otherwise, do not change your answer.
5) Now, summarize your planned bite from 4) in a list 
format. You can either output a single item in a list 
['item'], where 'item' is chosen from 'Items remaining', 
if you would like to feed this item by itself. Or, you 
can output two items in a list ['item1', 'item2'], if 
you would like to dip 'item1' (from 'Items remaining') 
in 'item2' (from 'Dipping sauces remaining') i.e. 
['banana', 'nutella']. +underline{However, you cannot suggest} 
+underline{to feed 2 items if one of them is not a dip (i.e.} 
+underline{['ramen', 'beef'] or ['rice', 'chicken'] or} 
+underline{['linguini', 'mushroom'] are NOT allowed, you must} 
+underline{feed ['ramen'] or ['beef'] by itself).} Lastly, you 
can output an empty list [] if there is absolutely no 
bite that makes sense.
---
Format your response as follows, be concise but 
thorough: Food Items Left: <Sentence describing what
food items are left on the plate; +underline{ensure that this} 
+underline{matches with the 'Items remaining' provided at the} 
+underline{beginning}>
Strategy: <Sentence describing your high-level 
strategy>
Next bite: <Phrase describing the next bite you plan to 
feed>
Next bite (accounting for efficiency): <Phrase 
describing the next bite you plan to feed, taking into
account efficiency>
Next bite as list: ['item1'] # Or ['item1', 'item2'] or 
[]
---
Output your response here. Ensure that the last line 
begins with 'Next bite as list:
\end{Verbatim}
\end{tcolorbox}

\begin{figure*}[t]
    \includegraphics[width=\linewidth]{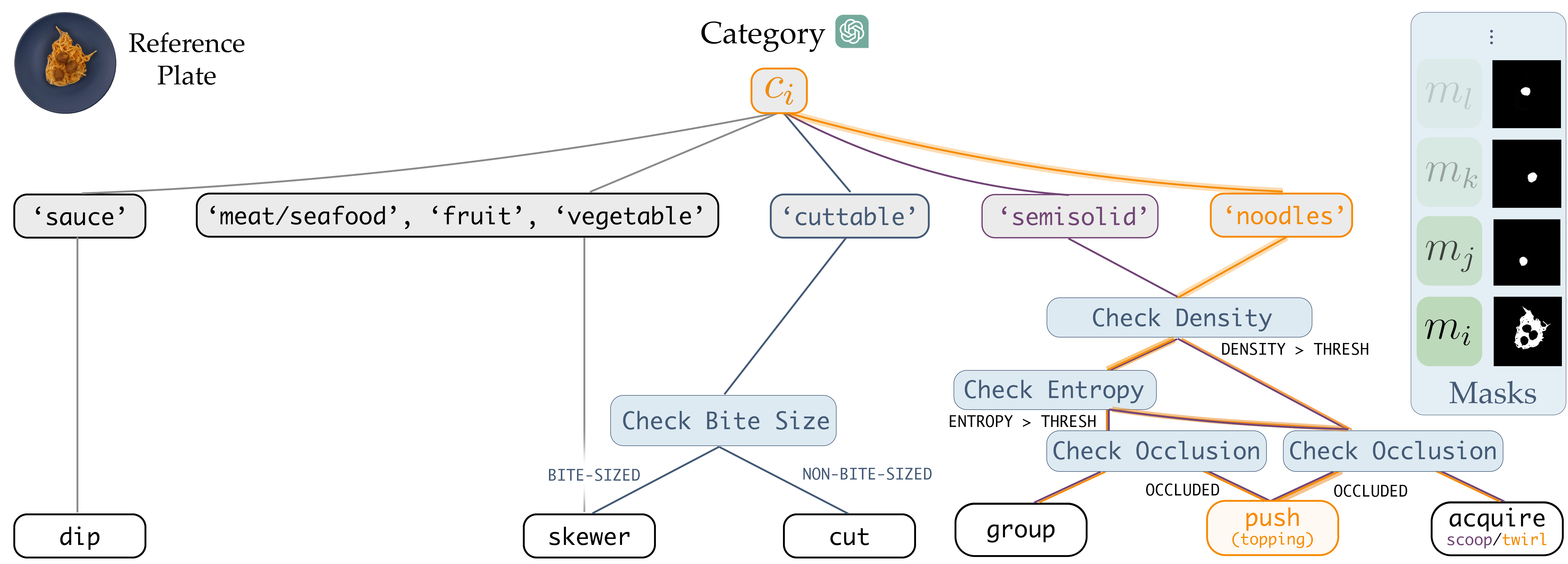}
    \caption{\textbf{Task Planner Decision Tree:} For this reference plate and input category \texttt{`noodles'}, the task planner navigates the decision tree to decide to push aside a meatball to access the bed of spaghetti underneath. Nodes in blue leverage the segmented food item states to determine branching.}
    \label{fig:task_planner_tree}
    \vspace{-0.3cm}
\end{figure*}

\section{Skill Parameterization Details}

At the core of \flair is a large library of vision-parameterized skills for both acquisition and pre-acquisition. We visualize each skill parameterization in \cref{fig:parameterization}.

The \texttt{twirl} skill and the end position for the \texttt{scoop} and \texttt{group} skills are parameterized by the densest point in a pile of noodles or semisolid mass. To sense the densest point, we take a mask $m_t$, apply a 2D Gaussian filter over the image, and consider the densest pixel to be the argmax of this heatmap. The \texttt{twirl} skill is additionally parameterized by the main-axis orientation of the pile of noodles at the densest point. To determine this, we use the pose estimation network from ~\cite{sundaresan2023learning} which maps a local crop of noodles to the predicted main axis orientation, and insert the fork orthogonal to this. Parameterizing the skill in this way encourages strands of noodles to be perpendicular and wrapped around the fork tines during twirling, rather than parallel and slipping off. The start positions for the \texttt{scoop} and \texttt{group} skills are parameterized by the sparsest point of a noodle or semisolid mask. To estimate this, we select the pixel on the segmentation mask that is furthest from the sensed densest point. For noodle or semisolid piles containing toppings, we update this parameterization to take into account the position of the filings, such that the downstream skill execution is not affected by the toppings. 

We parameterize the \texttt{dip} and \texttt{skewer} skills by the centroid of a predicted segmentation mask. The \texttt{skewer} skill is additionally parameterized by the mask's major-axis orientation, such that the tines of the fork run orthogonal to the item. The \texttt{push} skill starts at a constant offset from the centroid of the obstructing topping and pushes it to the nearest boundary point of the underlying mask. The fork tines are oriented perpendicular to the direction of the pushing action. 

For the \texttt{cut} skill, we aim to locate a bite-sized portion from an item. We first estimate the long axis of a food item via principal component analysis, and then compute the extremities of the segmentation mask. Starting from one extremity, we take the point which is a distance of a pre-defined constant $\texttt{BITE\_LENGTH}$ pixels away from the extremity point, along the direction of the main axis, and cut the food item orthogonal to the main axis at this point. 

\section{Task Planner Details}
To better understand the task planner $\tau$, we provide a visualization of the decision tree for different foods in \cref{fig:task_planner_tree}.


\newmdenv[backgroundcolor=blue!5]{boxedsection}

\section{Details on User Preferences}
\label{sec:user_prefs}
In this work, we consider user preferences that describe a desired \emph{ordering of bites}, and their \emph{combination with sauces}. While there are many other potential preferences to consider during feeding, such as combinations of multiple available bites, or desired bite sizes, our skill library is only equipped to handle one food item at a time (apart from dipping), without customizing for bite size. Thus, when soliciting preferences from user study participants, we use the following form template to provide context on preferences in scope with~\flair:

\begin{boxedsection}
    In this study, envision yourself receiving assistance from a robot caregiver tasked with feeding you. The robot is programmed to pick up and feed you one bite at a time from a plate filled with a variety of food items. The robot can only pick up one type of food item per bite. At the beginning of your meal, the robot asks for your input on the order in which you would like these items to be fed to you.

You might have specific preferences for the sequence in which the items are presented, such as starting with a particular item and ending with another, or alternating between items to enjoy a variety of flavors. These preferences can also vary in intensity. For instance, with a plate containing asparagus and mushrooms, you might express:

\begin{itemize}
    \item No particular preference: "I have no specific order preference for asparagus or mushrooms."
    \item No particular preference: "I'm fine with eating them in a random order, no specific pattern."
    \item Weak preference: "I sometimes begin with asparagus, but I'm not too strict about the order afterward."
    \item Weak preference: "I'd slightly prefer to eat all the asparagus first, but it's not crucial."
    \item Strong preference: "I always want to alternate between bites of asparagus and mushroom."
    \item Strong preference: "Please don't feed me any mushrooms."
\end{itemize}

Throughout this study, we will present you with eight different plates, each containing a unique assortment of food items. For each plate, we ask you to specify your preferred order for consuming these items. Please keep the following things in mind while responding - 

\begin{enumerate}
    \item Please write your preference in natural language, as if you are instructing your caregiver. 
If you have don't have a preference, you can write "No preference."
    \item Our robot caregiver is designed to pick up only one type of food item at a time for each bite. This means that for a plate with, say, asparagus and mushrooms, the robot can serve you bites of either asparagus or mushroom separately, but it's not able to combine them into a single bite.
    \item In this study, we're keen to learn about your preferences in terms of the order in which you'd like to be served different food items, and if applicable, your choices for dipping sauces for certain dishes. However, we kindly ask that you do not specify preferences related to the size of the bites or requests for cutting specific food items. 
\end{enumerate}

Your responses will help us understand the various ordering preferences people might have in a setting where they are fed by a robotic caregiver.
\end{boxedsection}

\vspace{-0.05cm}
As a representative sample of the kinds of responses received, we provide all preferences received for the spaghetti and meatballs plate below. Notably, we see that both shared and diverse preferences emerge, and very few preferences are out of scope (highlighted in {\color{BrickRed} red}) for \flair.
\vspace{0.15cm}

\begin{itemize}
\item \emph{\color{CadetBlue}``No particular preference''}
\item \emph{\color{BrickRed}``Cut up the meatballs and eat it with a bite of the pasta.''}
\item \emph{\color{CadetBlue}``Slight preference to twirl spaghetti around and to eat meatballs individually''}
\item \emph{\color{CadetBlue}``Would prefer to eat spaghetti only, without any meatballs.''}
\item \emph{\color{CadetBlue}``I would alternate between one bite of spaghetti and a meatball''}
\item \emph{\color{CadetBlue}``Id probably prefer eating a bite of spaghetti first before going in for the meat ball, but not a strong preference.''}
\item \emph{\color{CadetBlue}``Please feed me only the spaghetti, and skip all the metballs.''}
\item \emph{\color{CadetBlue}``No preference''}
\item \emph{\color{CadetBlue}``Spaghetti then meatballs''}
\item \emph{\color{CadetBlue}``alternating between spaghetti and meatballs''}
\item \emph{\color{CadetBlue}``I have no preference as long as the three meatballs are not fed consecutively.''}
\item \emph{\color{CadetBlue}``Eat fork of spaghetti then fork of meatball''}
\item \emph{\color{CadetBlue}``I would like to eat bites of spaghetti and meatballs in alternating order.''}
\item \emph{\color{CadetBlue}``I always want to alternate between bites of meatball and pasta.''}
\item \emph{\color{CadetBlue}``I'd slightly prefer to alternate a bit between the spaghetti and meatballs.''}
\item \emph{\color{CadetBlue}``I prefer to eat meatballs first but it's not crucial.''}
\item \emph{\color{CadetBlue}``I like to alternate bites: one bite spaghetti followed by one bite of meatball, and so on.''}
\item \emph{\color{CadetBlue}``First, roll up spaghetti onto fork and eat. Part and then stab the meatball with the fork and eat. I would like to have at least one fork of spaghetti before and in between the meatballs.''}
\item \emph{\color{CadetBlue}``I always alternate between bites of spaghetti and meatballs.''}
\item \emph{\color{CadetBlue}``I always want to alternate between bites of meatballs and spaghetti.''}
\item \emph{\color{CadetBlue}``I'd always like to alternate between spaghetti and meatballs, starting with spaghetti first.''}
\item \emph{\color{CadetBlue}``Eat spagetti first, then any order would work''}
\item \emph{\color{CadetBlue}``Please don't feed me meatballs.''}
\item \emph{\color{CadetBlue}``I would like to eat all of the spaghetti. Please don't feed me any meatballs.''}
\item \emph{\color{CadetBlue}``I prefer eating more spaghetti at first followed by randomly switching between spaghetti and meatballs.''}
\item \emph{\color{CadetBlue}``Any order is fine in the beginning as long as the last 2 bites are spaghetti''}
\item \emph{\color{CadetBlue}``Id prefer to alternate between bites of spaghetti and meatballs, but it is not crucial,''}
\item \emph{\color{CadetBlue}``I would get a few bites of spaghetti, then one meatball, then alternate so that I finish them at approximately the same time.''}
\item \emph{\color{CadetBlue}``I would prefer to alternate between bites of spaghetti and bites of the meatballs.''}

\begin{figure*}[t]
    \includegraphics[width=\linewidth]{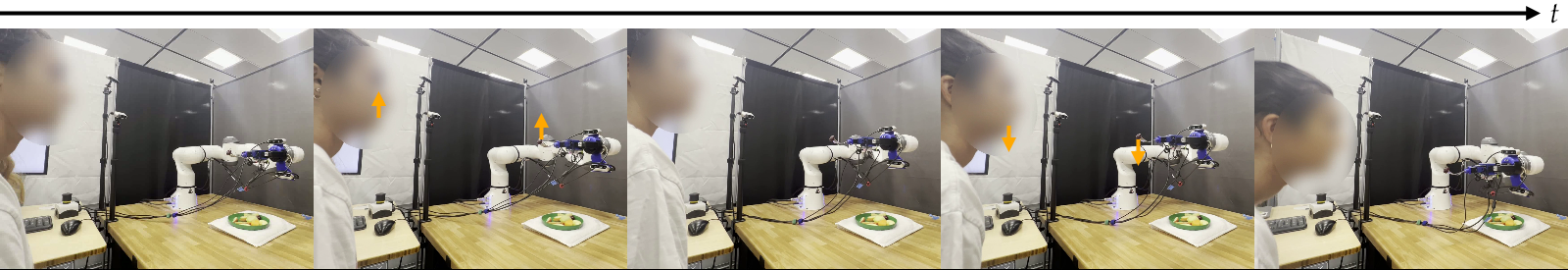}
    \caption{\textbf{Bite Transfer via Visual Servoing:} \flair~ is compatible with any method of bite transfer, and we demonstrate this by combining acquisition with a visual-servoing based approach on the Franka Panda. We use an open-source \href{https://github.com/1adrianb/face-alignment}{facial landmark tracking library} to dynamically track a person's mouth position, and update the position of the fork-tip such that transfer is robust to small head movements (orange arrows).}
    \label{fig:servo}
\end{figure*}

\item \emph{\color{CadetBlue}``Alternate between spaghetti and meatballs.''}
\item \emph{\color{CadetBlue}``I sometimes prefer to alternate between spaghetti and meatballs''}
\item \emph{\color{CadetBlue}``Please feed me meatballs at regular intervals of time so that I consistently get bites of meatballs until they run out. Between bites of meatballs, feed me spaghetti. Multiple bites of spaghetti between bites of meatballs is fine. The goal is that I regularly get the meatballs and they do not run out by the end. This is not a strong requirement but it would be great if last bite is of the meatballs.''}
\item \emph{\color{CadetBlue}``Twirl a fork for spaghetti, stab meatballs with fork''}
\item \emph{\color{BrickRed}``I would alternate between cutting a meatball in half, skewering it, then eating it and twirling spaghetti onto my fork then eating it. I would repeat this process till the plate is empty.''}
\item \emph{\color{CadetBlue}``I would prefer to eat a meat ball first.''}
\item \emph{\color{CadetBlue}``I like to eat spaghetti first and then afterwards eat the meatballs.''}
\item \emph{\color{CadetBlue}``I'd like to alternate between meatballs and spaghetti but it isn't crucial''}
\item \emph{\color{CadetBlue}``I would alternate between bites of spaghetti and bites of meatball''}
\item \emph{\color{CadetBlue}``I only eat the meatballs.''}
\item \emph{\color{CadetBlue}``Alternating bites''}
\item \emph{\color{CadetBlue}``I have a weak preference for alternating spaghetti and meatballs, maybe a couple bites of spaghetti followed by a meatball.''}
\item \emph{\color{CadetBlue}``I always want to alternate between bites of spaghetti and meatball.''}
\item \emph{\color{CadetBlue}``Please don't feed me meatballs.''}
\item \emph{\color{CadetBlue}``I'd prefer to first eat some spaghetti with meatballs in middle. Depending on how much I like the meatballs, I would decide if I want my last bite to be spaghetti or meatballs.''}
\item \emph{\color{CadetBlue}``Please dont feed me any meatballs. I would take alternating bits of spaghetti and meatballs''}
\item \emph{\color{CadetBlue}``I’d like to have the spaghetti first and then a meatball, and then repeat the order''}
\item \emph{\color{CadetBlue}``Do not feed me meatballs''}
\item \emph{\color{CadetBlue}``No particular preference''}
\item \emph{\color{CadetBlue}``Starting with meatball, followed by alternation of meatballs and spaghetti, and with spaghetti''}
\end{itemize}

From these responses, we then use LLM summarization to find groups of common preferences (i.e. \emph{\color{CadetBlue}``Alternate bites of spaghetti and meatballs''} or \emph{\color{CadetBlue}``Don't feed me meatballs, only spaghetti''}) with which to evaluate.

\section{Additional Experimental Results}
\subsection{Efficiency Across Methods} 
\begin{figure}[t]
    \centering
    \vspace{-10pt}
    \includegraphics[width=0.4\textwidth]{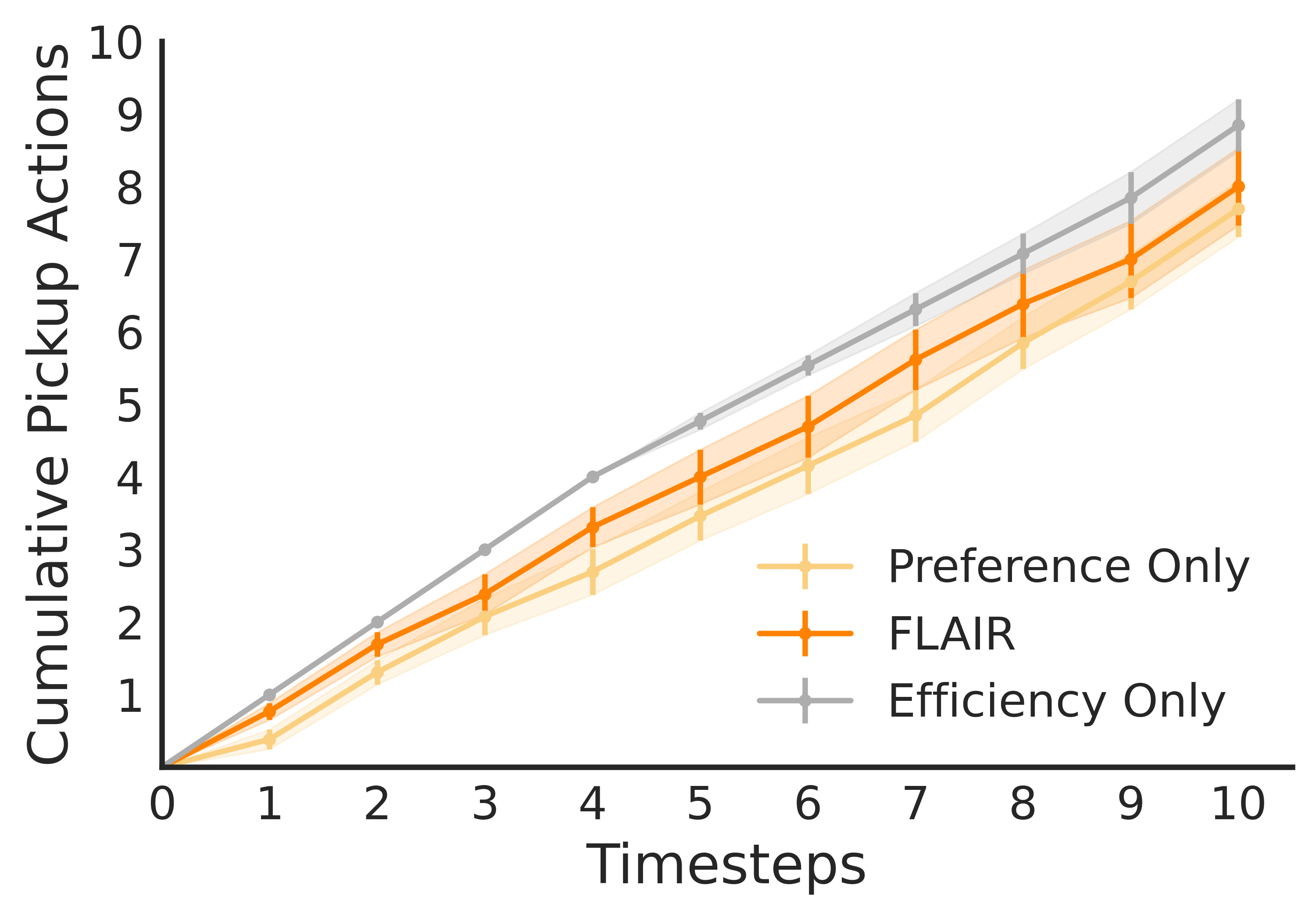}
    \caption{\textbf{Efficiency Across All Plates:} Across all evaluated plates, including plates where both strong preferences or no preferences were provided, we find that \flair~ outperforms Preference/Commonsense Only in terms of amount of food picked up over time. }
    \label{fig:all_prefs_efficiency}
    \vspace{-13pt}
\end{figure}

We empirically find that the trend of \flair outperforming a Commonsense/Preference Only approach in terms of efficiency  holds both in the presence and absence of strong user preferences. In Figure \ref{fig:all_prefs_efficiency}, we visualize the amount of food picked up over time across all methods, averaged across \emph{all} trials including strong user preferences and none at all. We see that \flair yields more pickup actions over time compared to Preference Only. It is unsurprising that Efficiency Only performs the best, but this is expected as \flair must also balance actions that satisfy given preferences as well, when a user has strong preferences.

\subsection{Bite Transfer} 
In addition to integrating \flair with both inside-mouth and outside-mouth bite transfer frameworks on the Kinovas, we additionally show that \flair is compatible with any bite transfer framework by pairing it with a visual servoing-based outside-mouth framework on the Franka. In \cref{fig:servo}, we provide qualitative visualizations showing the ability of the system to react to head movements in real-time, and adjust the fork tip position accordingly to align with the mouth's center. 

\end{document}